\documentclass{article}
\usepackage{PRIMEarxiv}
\usepackage[utf8]{inputenc} 
\usepackage[T1]{fontenc}    
\usepackage{url}            
\usepackage{booktabs}       
\usepackage{amsfonts}       
\usepackage{nicefrac}       
\usepackage{microtype}      
\usepackage{lipsum}
\usepackage{graphicx}       
\usepackage{listings}
\usepackage{cancel}
\usepackage{soul}
\usepackage{makecell}
\usepackage{amsmath}
\usepackage{tabu}
\usepackage{tikz}
\usepackage{mathdots}
\usepackage{yhmath}
\usepackage{cancel}
\usepackage{color}
\usepackage{siunitx}
\usepackage{array}
\usepackage{multirow}
\usepackage{amssymb}
\usepackage{tabularx}
\usepackage{extarrows}
\usepackage{booktabs}
\usepackage{bbm}
\usetikzlibrary{fadings}
\usetikzlibrary{patterns}
\usetikzlibrary{shadows.blur}
\usetikzlibrary{shapes}
\usepackage{amsmath}
\usepackage{amsfonts}
\usepackage{algorithmic}
\usepackage{algorithm2e}
\SetKwInput{KwInput}{Input}
\SetKwInput{KwOutput}{Output}
\RestyleAlgo{ruled}
\usepackage{setspace}

\usepackage{natbib}
\usepackage{graphicx}
\usepackage[colorlinks=true, allcolors=blue]{hyperref}
\usepackage{tabularx}
\usepackage{booktabs,siunitx,array,threeparttable}
\usepackage{soul}

\usepackage{enumitem,amssymb}
\newlist{todolist}{itemize}{2}
\setlist[todolist]{label=$\square$}
\usepackage{pifont}

\usepackage{lipsum} 
\usepackage{fancyhdr}
\title{Liner Shipping Network Design with Reinforcement Learning
}

\author{
  Utsav Dutta, Yifan Lin, Zhaoyang Larry Jin \\
  Data Science \\
  C3 AI \\
  Redwood City, California, US\\
  \texttt{\{utsav.dutta, yifan.lin, larry.jin\}@c3.ai} \\
}

\begin{document}
\maketitle

\begin{abstract}
This paper proposes a novel reinforcement learning framework to address the Liner Shipping Network Design Problem (LSNDP), a challenging combinatorial optimization problem focused on designing cost-efficient maritime shipping routes. Traditional methods for solving the LSNDP typically involve decomposing the problem into sub-problems, such as network design and multi-commodity flow, which are then tackled using approximate heuristics or large neighborhood search (LNS) techniques. In contrast, our approach employs a model-free reinforcement learning algorithm on the network design, integrated with a heuristic-based multi-commodity flow solver, to produce competitive results on the publicly available LINERLIB benchmark. Additionally, our method also demonstrates generalization capabilities by producing competitive solutions on the benchmark instances after training on perturbed instances.

\end{abstract}




  







\section{Introduction}\label{sec:intro}

The liner shipping industry is the backbone of global maritime trade. It plays a critical role in the international supply chain, ensuring the efficient movement of merchandise across the globe. This industry involves the design and operation of container vessels that traverse fixed maritime routes to transport goods between ports efficiently and profitably. The strategic planning of these routes plays a pivotal role in optimizing both the revenue of ocean freight companies and the operational efficiency of their vessels. Well-designed shipping networks not only enhance profitability but also reduce the total number of vessels utilized, leading to lower maintenance costs and decreased emissions.

The Liner Shipping Network Design Problem (LSNDP) addresses this complex business challenge by modeling this as a mathematical optimization problem. The goal of the LSNDP is to design optimal vessel routes and allocate cargo flows across the network to maximize overall profitability. As a specialized routing problem, the LSNDP falls under the broader category of combinatorial optimization problems. Similar routing problems include the well-known Traveling Salesman Problem (TSP) and Vehicle Routing Problem (VRP). Similar to these, the LSNDP is classified as NP-hard, due to which solving large-scale instances to optimality with traditional OR approaches such as Mixed-Integer Programming (MIP) is computationally intractable.

However, recent advances in deep learning for routing problems have introduced an alternative approach to solving NP-hard combinatorial optimization problems. \cite{kool2018attention} applied Reinforcement Learning (RL) to the Traveling Salesman Problem (TSP) and several variants of the Vehicle Routing Problem (VRP), demonstrating the potential of RL in this domain. Building on this, \cite{joshi2019efficient} enhanced the RL framework by incorporating Graph Convolutional Networks (GCN), which yielded promising results for TSP. These learning-based approaches achieved results comparable to traditional OR methods in terms of optimality, while also demonstrating impressive generalizability. This opens up avenues to learn general purpose policies from a diverse dataset and use these to infer high quality solutions on new unseen data points.

In this paper, we present a learning-based approach to the LSNDP. We break down the LSNDP into two sub-problems, as is classically done in OR-based approaches to this problem, namely the Network Design Problem (NDP) and the Multi Commodity Flow Problem (MCF). We formulate the NDP as a Markov Decision Process (MDP). By integrating a reinforcement learning (RL) algorithm with a heuristic-based multi-commodity flow solver, our method achieves competitive results on the publicly available LINERLIB \footnotemark[1] benchmark. Our approach offers its value in two distinct ways. Firstly, our approach demonstrates that it can serve as a competitive end to end optimizer, in a similar vein to traditional OR solvers, and secondly, our approach shows signs of generalization capabilities by learning a robust policy that can provide high quality solutions for perturbed instances.

\footnotetext[1]{https://github.com/blof/LINERLIB}

The rest of the paper is organized as follows. We introduce the Liner Shipping Network Design Problem (LSNDP) and its decomposed sub-problems: multi-commodity flow (MCF) and network design problem (NDP) in Section~\ref{sec:lsndp}. For the decomposed network design problem, we introduct our reinforcement learning framework in Section~\ref{sec:policy_network}. We demonstrate the quality and generalizability of the proposed approach on the LINERLIB benchmark in Section~\ref{sec:result}. Finally, we conclude the paper in Section~\ref{sec:conclusion}.

\section{Related Work}\label{sec:relatedwork}

The Liner Shipping Network Design Problem (LSNDP) has been extensively researched within the operations research (OR) community for several decades. To support benchmarking efforts for the LSNDP, \cite{brouer2014base} introduced a standardized dataset known as LINERLIB\footnotemark[1], which includes seven real-world instances of the problem, each varying in scale. In a comprehensive review of the LSNDP literature, \cite{christiansen2020liner} discussed the standardized formulation of the problem widely accepted by the OR community and reviewed the development of the OR-based approaches typically used in this domain. The performance of leading algorithms is benchmarked on the LINERLIB dataset in their work. According to the authors, traditional OR approaches to the LSNDP can generally be categorized into the following main types:
\begin{itemize} 
    \item Holistic MIP-based formulations, which address both the service design and the multi-commodity flow aspects simultaneously, as exemplified by the work of \cite{plum2014service} and \cite{wang2014liner}.
    \item Local search-based methods, which explore variations from a predefined set of candidate services, such as the approach described by \cite{meng2011liner} and \cite{balakrishnan2017container}.
    \item Two-stage algorithms: These decompose the problem into two distinct phases: first solving the network design problem (NDP) and then the multi-commodity flow (MCF) problem separately. A common method, as used by \cite{brouer2014base} and \cite{thun2017analyzing}, involves designing the services (i.e., NDP) first and subsequently routing the containers through the designed network (i.e., MCF). Alternatively, \cite{krogsgaard2018flow} propose a reverse approach, where containers are first flowed through a relaxed network before finalizing the network design based on the flow. This class of approaches has generally proven to be the most effective, yielding reasonable solutions on the largest instance in the LINERLIB dataset, where other methods have failed.
\end{itemize}
Despite these advancements, two significant challenges persist across all OR methods. First, scalability remains a critical issue due to the NP-hard nature of the LSNDP, with large, real-world instances still unsolved. Second, generalizability is a major limitation, as even small perturbations in problem instances often necessitate a complete reconstruction of the solution, requiring a similar level of computational effort as the original problem. 

In the past decade, Reinforcement Learning (RL) has gained increasing attention as a method for solving combinatorial optimization problems. \cite{khalil2017learning} pioneered the use of RL to tackle the Maximum Cut and Minimum Vertex Cover problems, combining graph embeddings with Q-learning to generate heuristics. \cite{hu2017solving} made the first attempt at applying RL to the Bin Packing Problem. Within the domain of routing problems, \cite{vinyals2015pointer} introduced the Pointer Network (PN) to address the Traveling Salesman Problem (TSP), employing attention mechanisms to map inputs to outputs. \cite{bello2016neural} built on this work by applying the Actor-Critic algorithm to improve PN performance. \cite{nazari2018reinforcement} extended the RL approach to the Vehicle Routing Problem (VRP), enhancing the Pointer Network by replacing the Long-Short Term Memory (LSTM) encoder with a 1-D convolutional embedding.

More recent advances in RL for routing problems include the work of \cite{kool2018attention}, who developed a construction-heuristic learning approach applied to the TSP, VRP, and other related routing challenges. Their approach enhanced the encoder by introducing a Transformer-like attention mechanism, while the decoder maintained a similar structure to the original PN. \cite{joshi2019efficient} further advanced the field by incorporating a Graph Convolutional Network (GCN) for the encoder, alongside a highly parallelized, non-autoregressive beam search roll-out. This method outperformed autoregressive models in terms of solution quality for TSP. Building on these developments, \cite{fellek2023graph} introduced a more sophisticated graph embedding scheme, leveraging a multi-head attention structure that embeds edge information. Their approach demonstrated notable improvements for VRP, further advancing the effectiveness of RL in solving complex routing problems.

Significant strides have been made to improve the scalability and generalizability of RL-based approaches. \cite{drori2020learning} tackled edge-selection problems by converting them into node-selection tasks using line graphs and applied Graph Attention Networks (GAT) for embedding. Their method was tested on various NP-hard combinatorial optimization problems, including TSP, and showed that inference time scaled linearly with problem size. Other key advancements include the Adaptive Multi-Distribution Knowledge Distillation (AMDKD) framework proposed by \cite{bi2022learning}, and the work of \cite{fu2021generalize}, which employed Monte Carlo Tree Search (MCTS) alongside a heat map generated from a pre-trained supervised learning model to effectively scale solutions for arbitrarily large VRP instances.

The advancements of the above-mentioned RL methods show potential for application to a wider range of routing problems beyond TSP and VRP, including the Liner Shipping Network Design Problem (LSNDP). This paper attempts to take strides in that direction, by applying RL to a richer, and harder class of NP-hard combinatorial optimization problems, which has significant applications to real-world applications.

\section{LSNDP}\label{sec:lsndp}

\cite{brouer2014base} and \cite{christiansen2020liner} both provided a comprehensive definition of the Liner Shipping Network Design Problem (LSNDP): Given a set of ports, a fleet of container vessels, and a collection of demands specified in the quantity of Forty-Foot Equivalent units (FFE) with designated origins, destinations, and shipping rates, the objective is to design a set of cyclic sailing routes for the vessels (i.e., services) that maximize revenue from fulfilling the demands while minimizing the overall operational costs of those vessels.

It is important to note that, unlike the VRP or other dispatch problems, the LSNDP does not focus on the specific scheduling of vessels. Instead, it generates a set of services, each representing a round-trip route with a fixed itinerary of ports, called at regular intervals, typically at a weekly or bi-weekly frequency. Vessel assignments are subsequently determined to support these services. For example, if a round-trip route takes six weeks to complete, six vessels will be needed to maintain a weekly service. Additionally, the demand in LSNDP is normalized according to these service frequencies, simplifying the problem formulation.

\begin{figure}[htp]
    \centering
    \includegraphics[width=0.5\linewidth]{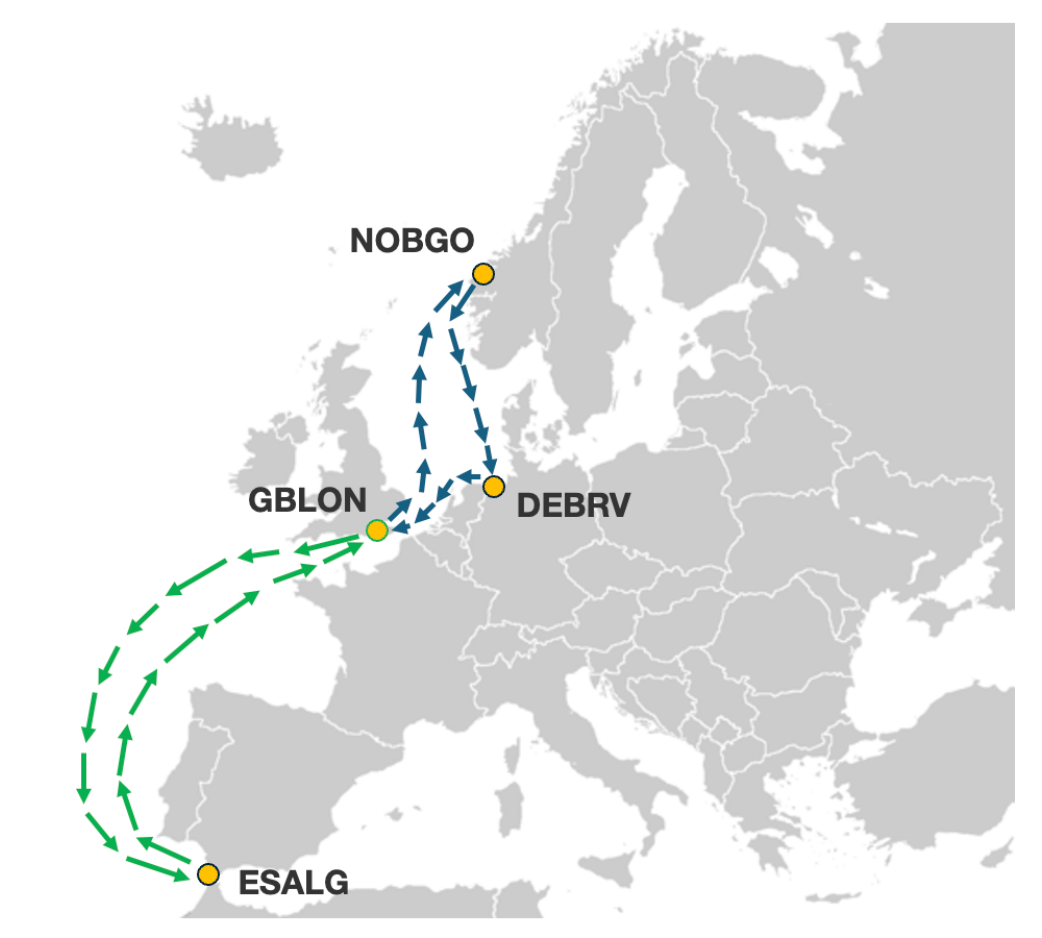}
    \caption{Example of a butterfly service with a hub port at London (GBLON).}
    \label{fig:service}
\end{figure}

Services in the LSNDP are categorized based on the topological structure of their routes. A simple service follows a round-trip route where vessels visit each port exactly once, forming a single circular loop. However, services are often non-simple, meaning that some ports are visited more than once along the same route. Visually, these non-simple services form multiple loops. Depending on their structure, they can be classified as butterfly services, pendulum services, or complex services. Figure~\ref{fig:service} illustrates an example of a butterfly service generated in the LSNDP, where ports are labeled using their corresponding UNLOCODE\footnotemark[2]. Note that one hub port London (GBLON) is visited twice.

\footnotetext[2]{https://unece.org/trade/uncefact/unlocode}

Building on the basic definition provided above, several variations of the LSNDP have been extensively studied in the literature. These variations introduce additional factors such as transit time, which imposes time constraints on demand, transshipment costs, vessel speed optimization, and penalties for leaving a portion of the demand unsatisfied (rejected demand). In line with most studies that benchmark their results using the LINERLIB dataset, this work focuses on the LSNDP variation that incorporates transshipment costs and rejected demand, while excluding considerations for transit time and vessel speed optimization. Additionally, we assume that vessels operate strictly at their designed speed and permit fractional vessel assignments, which simplifies the modeling of vessel deployment and optimizes resource allocation.

As discussed in Section~\ref{sec:relatedwork}, a widely adopted approach within the traditional operations research (OR) community for solving the Liner Shipping Network Design Problem (LSNDP) is to decompose it into two closely related sub-problems: the multi-commodity flow (MCF) problem and the network design problem (NDP). Detailed descriptions of the MCF and NDP can be found in Appendix~\ref{app:lsndp}.

In our proposed reinforcement learning (RL) approach, we aim to leverage this two-tier framework. Rather than formulating the NDP as a Mixed-Integer Problem (MIP), we represent it as a Markov Decision Process (MDP), where round-trip services are generated sequentially. At each step $t$, the generation of a complete service is treated as an action $A_t$, with the step index $t$ indicating the number of services generated up to that stage.

The MCF serves as a key component in the reward evaluation function at each step, defined by:
\begin{align}
    R_{t+1} = \eta_{t+1} - \eta_{t},\label{eq:reward}
\end{align}
where $R_{t+1}$ is the reward in the MDP context, and $\eta_{t+1}$ and $\eta_{t}$ represent the profit of the network with all services generated up to steps $t+1$ and $t$, respectively. For further details on the MDP formulation, refer to Appendix~\ref{app:mdpdetails}.

To support the RL approach, an advanced MCF algorithm is crucial for enabling fast reward evaluation, as the reward function is invoked hundreds of millions of times during the training process. Additionally, the quality of the NDP solution is partly influenced by the performance of the MCF algorithm since variations in the reward signals can steer the RL training in different directions. In this study, we have implemented a basic version of the MCF algorithm (see Appendix~\ref{app:heuristic_mcf}), which is less sophisticated than the state-of-the-art heuristic implementations, such as those by \cite{krogsgaard2018flow}, and also lags behind MIP-based solutions.

Despite the limitations of our MCF implementation, we were still able to develop an RL-based solution for the NDP component of the LSNDP. In the following section, we introduce two approaches for finding optimal NDP solutions by parameterizing the policy as neural networks.

\section{Policy Neural Network Design}\label{sec:policy_network}
In this section, we present two modeling approaches for parameterizing the policy $\pi_{\theta}$ as neural networks: the encoder-only approach and the encoder-decoder approach. After representing the NDP as a Markov Decision Process (MDP), either approach can produce a parameterized policy $\pi_{\theta}$ that guides the actions at each step $t$. This process can be described by the following sampling equation:
\begin{align}
    A_{t} \sim \pi_{\theta}(\cdot | S_{t}),
\end{align}
where the action $A_{t}$ in the context of the NDP comprises of two components: the vessel selection $A_{v,t} \in \mathbb{R}^{V}$, which determines the vessel type to be deployed from the overall fleet, and the service selection $A_{p,t} \in \mathbb{R}^{P}$, which specifies an ordered sequence of ports:
\begin{align}
    A_{t} = [A_{v, t}, A_{p, t}]. \label{eq:action}
\end{align}
The state at step $t$ is represented by two components: $S_{t} = \{S_{t,g}, S_{t,v} \}$. Here, $S_{t,g}$ captures the state of the shipping network as a graph, and $S_{t,v}$ describes the status of the available vessels. These components are defined as follows:
\begin{align}
    S_{t,g} & =  \{\mathbf{f}_p,\mathbf{f}_e\}, \label{eq:s_g}\\
    S_{t,v} & = \mathbf{v}_t \in \mathbb{R}^{V\times D_v}. \label{eq:s_v}
\end{align}
In this representation, $\mathbf{f}_p \in \mathbb{R}^{(P+1)\times 2}$ contains the two-dimensional features of all nodes in the graph, where each node corresponds to a port in the shipping network. The notation $P$ represents the total number of ports in a given problem instance, while the two-dimensional features capture the total incoming and outgoing demand for each port at step $t$. An additional node is included to track the global features of the graph, resulting in a total of $P+1$ nodes. The feature matrix $\mathbf{f}_e \in \mathbb{R}^{E \times D_e}$ describes the edges in the graph, where each edge represents a potential leg in a service connecting two ports. Here, $E$ represents the total number of possible port pairs that a vessel can traverse in a single leg, and $D_e$ is the dimension of the edge features, with $D_e=6+|S|$, where $|S|$ is the maximum number of services allowed in the problem instance. Note that features $\mathbf{f}_e$ and $\mathbf{f}_p$ are step $t$ dependent, here we drop the subscript for simplicity.

The notation $\mathbf{v}_t$ represents the feature set for each vessel class, including the number of vessels remaining to be deployed in each class. Here, $V$ denotes the total number of vessel classes in the problem instance, and $D_v$ represents the dimension of the vessel features. For more details on the state representations, please refer to Appendix~\ref{app:mdpdetails}.

Both the encoder-only and encoder-decoder approaches follow a similar sequence: selecting the vessel class first, and then determining the service rotation. The sequence repeats at each step $t$ until one of the following conditions is met: the maximum number of services $|S|$ is reached, all demands are satisfied, or all vessels are exhausted. The key difference lies in how the service rotation $A_{p,t}$ is generated. Note, $A_{p,t}$ represents the selection of one service for the shipping network, which involves determining a sequence of ports to include in the service. This can be approached in two ways: either selecting all ports simultaneously and deciding their sequence afterward (encoder-only approach) or selecting each port sequentially (encoder-decoder approach).

The encoder-only approach uses a one-shot rollout, where a complete sequence of actions is generated in a single step, based on a policy that predicts the entire sequence simultaneously. In contrast, the encoder-decoder approach relies on an autoregressive rollout, where the policy generates ports sequentially one at a time, using each previously selected port as input for the next selection. The complete sequence of ports once generated constitutes an action.

\subsection{Encoder-Only Approach with One-Shot Rollout}\label{subsec:encoder}
\begin{figure}[htp]
    \centering
    \includegraphics[width=0.9\textwidth]{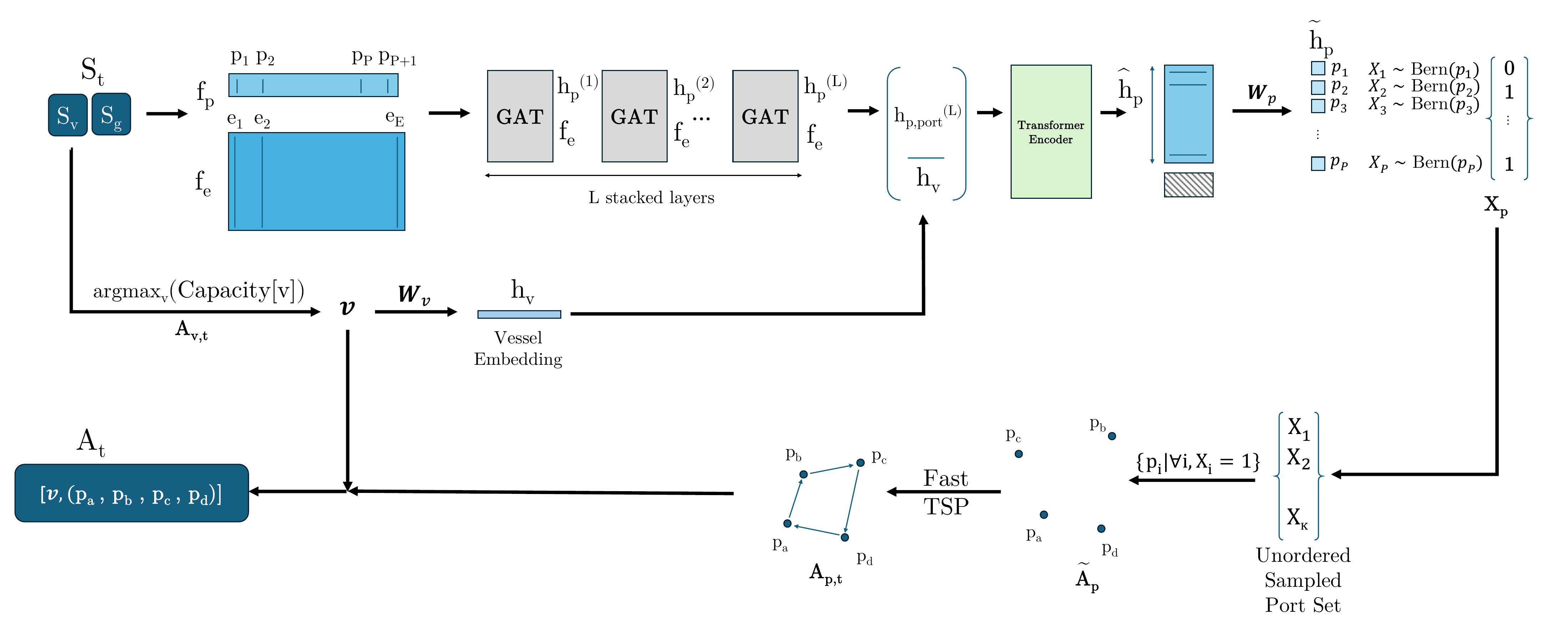}
    \caption{Encoder policy diagram for NDP.}
    \label{fig:encoder}
\end{figure}

Figure~\ref{fig:encoder} demonstrates the workflow of the encoder-only approach. Instead of defining vessel selection as sampling from a stochastic policy, we use a simple deterministic heuristic for vessel selection. Specifically, we select the ``largest available vessel'' as the action $A_{v,t}$, where ``largest'' refers to the vessel class with the highest capacity, measured in FFEs:
\begin{align}
    A_{v,t} = \operatorname*{argmax}_{\substack{ v \in [1, V] \\ (\mathbf{v}_t)_{v,1}>0}}\text{Capacity}(v). \label{eq:encoder_vessel_selection}
\end{align}
Here, $v$ represents the index of the vector $\mathbf{v}_{t}$, and Capacity($v$) denotes the capacity of the vessel class corresponding to index $v$. The constraint $(\mathbf{v}_t)_{v,1} > 0$ ensures that only vessel classes with available vessels are considered for selection.

Our encoder is composed of $L$ sequential layers of Graph Attention Networks (\textbf{GAT}) with an embedding dimension of $H$, followed by a standard Transformer encoding layer. The GAT layers learn contextual representations for each port by leveraging the underlying graph structure, while the Transformer encoder enables attention across all port pairs, capturing interactions between them. We omitted the positional embedding layer, as our embedding structure does not have inherent temporal properties.

We pass the graph representation, $\mathbf{f}_p$ and $\mathbf{f}_e$, through the GAT layers. After processing it through $L$ graph attention layers, we obtain a dense, contextualized representation $\mathbf{h}^{(L)}_{p} \in \mathbb{R}^{(P+1)\times H}$. Mathematically, this process can be expressed as follows:
\begin{align}
    \mathbf{h}^{(1)}_{p} & =  \textbf{GAT}^{(1)}(\mathbf{f}_p,\mathbf{f}_e) \in \mathbb{R}^{(P+1)\times H}, \\
    \mathbf{h}^{(l)}_{p} & =  \textbf{GAT}^{(l)}(\mathbf{h}^{(l-1)}_{p},\mathbf{f}_e) \in \mathbb{R}^{(P+1)\times H},
\end{align}
where the superscript $(l)$ denotes the $l$-th layer of the network. Note that only the node features are transformed into the dense representation $\mathbf{h}^{(l)}_{p}$, while the edge features $\mathbf{f}_e$ remain unchanged. Through this iterative message-passing process, information from both the nodes and edges is effectively aggregated into the node representations.

It is important to note that the output from the GAT consists of two components: the port embeddings and the global embedding:
\begin{align}\label{eq:encoder_embedding}
    \mathbf{h}_{p}^{(L)} = [\mathbf{h}_{p, \text{port}}^{(L)}, \mathbf{h}_{p, \text{global}}^{(L)}] \in \mathbb{R}^{(P+1)\times H},
\end{align}
where $\mathbf{h}_{p, \text{port}}^{(L)} \in \mathbb{R}^{P\times H}$ represents the embeddings of the ports, and $\mathbf{h}_{p, \text{global}}^{(L)} \in \mathbb{R}^{1\times H}$ corresponds to a global node embedding that serves as a general representation of the entire graph. In the encoder-only approach, only the port embeddings are utilized, while the global embedding is used by the decoder.

With the port information fully encoded, we now shift our focus to the vessel information. Based on the vessel action $A_{v,t}$ described earlier, $v$ denotes the index of the selected vessel class. In this step, only the features of the selected vessel class, $(\mathbf{v}_{t})_v \in \mathbb{R}^{D_v}$, are used in the workflow. The vessel state is then encoded using a matrix multiplication:
\begin{align}
    \mathbf{h}_{v} = \mathbf{W}_{v}(\mathbf{v}_{t})_v \in \mathbb{R}^{H}, \label{eq:encoder_vessel_embed}
\end{align}
where $\mathbf{W}_{v} \in \mathbb{R}^{H\times D_v}$ is a linear transformation that maps the vessel state features to a dense representation of dimension $H$.

After encoding both the graph state and vessel state into dense matrices, they are processed through a standard Transformer encoder. The resulting updated graph embedding $\hat{\mathbf{h}}_{p}$ is then passed through a Sigmoid function:
\begin{align}
    [\hat{\mathbf{h}}_{p}, \hat{\mathbf{h}}_{v}] & = \textbf{Transformer}(\mathbf{h}_{p, \text{port}}^{(L)}, \mathbf{h}_{v}) \in \mathbb{R}^{(P+1)\times H}, \label{eq:encoder_transformer} \\
    \tilde{\mathbf{h}}_{p} & = \frac{1}{1+ e^{-(\mathbf{W}_{p}\hat{\mathbf{h}}_{p})^{T}}} \in \mathbb{R}^{P}, 
\end{align}
where $\mathbf{W}_p$ is a linear transformation that maps the graph embedding to a $P$-dimensional vector. The Sigmoid function produces $\tilde{\mathbf{h}}_{p} \in \mathbb{R}^{P}$, with all elements constrained within the range $[0,1]$. The resulting embedding is then subjected to a Bernoulli sampling process, defined by:
\begin{align}
    \mathbf{X}_{p} = \textbf{Bernoulli}(\tilde{\mathbf{h}}_{p}) \in \{0,1\}^{P}.
\end{align}
We define the set $\tilde{A}_p=\{i| (\mathbf{X}_p)_i = 1, \forall i \in \mathbf{X}_p\}$, which represents an unordered set of ports to be included in a service. A port is included in the service if the corresponding value in $\mathbf{X}_p$ is 1, and excluded if the value is 0.

To generate an ordered set representing a service, we use a fast approximate TSP solver (see \cite{shintyakov2017tsp}) to transform the unordered set into an ordered rotation:
\begin{align}
    A_{p,t} = \textbf{TSP}(\tilde{A}_{p}, \mathbf{f}_e). \label{eq:encoder_tsp}
\end{align}
The TSP solver requires additional static graph features from $\mathbf{f}_e$ (such as the distance matrix between ports) to determine the optimal port-call sequence. The resulting ordered set of ports is treated as the service selection action for the current step, $A_{p,t}$. Here, we include the time step subscript $t$ to maintain consistency with the notation used in other sections. Assembling $A_{v,t}$ from Eq.~\ref{eq:encoder_vessel_selection} and $A_{p,t}$ from Eq.~\ref{eq:encoder_tsp}, action $A_t$ is completed as defined in Eq.~\ref{eq:action}.

\subsection{Encoder-Decoder Approach with Autoregressive Rollout}\label{subsec:enc_dec}
Equations~\ref{eq:encoder_transformer} to \ref{eq:encoder_tsp} describe the one-shot rollout for the encoder-only approach, where the probability of each port being included in a service is modeled independently within each action $A_t$. While this method is straightforward and intuitive, it treats the inclusion of each port as independent, limiting its ability to account for dependencies between selected ports. In contrast, the encoder-decoder approach with autoregressive rollout, explicitly models these dependencies, where the decision to include an additional port depends on the previously selected ports in the current and all previous services.

In the autoregressive rollout, the action $A_t$ is generated sequentially, involving multiple sub-steps within a single step $t$. The process begins with a sub-step, denoted as $\tau$, for vessel selection, followed by several sub-steps to select ports, thereby completing the generation of a single service. It is important to note that in this approach, vessel selection is also determined by the policy $\pi_{\theta}$, which is parameterized by a neural network, rather than the rule-based selection used in the encoder-only approach (Eq.~\ref{eq:encoder_vessel_selection}).

The embeddings used for the autoregressive rollout are generated as outputs from the encoder phase. Specifically, for the port embeddings, we define:
\begin{align}
    \check{\mathbf{h}}_{p} = \textbf{Transformer}(\mathbf{h}_{p, \text{port}}^{(L)}) \in \mathbb{R}^{P \times H},
\end{align}
where $\mathbf{h}_{p, \text{port}}^{(L)} \in \mathbb{R}^{P\times H}$ is the port embedding previously defined in Eq.~\ref{eq:encoder_embedding}. For the vessel embeddings, we similarly define:
\begin{align}
    \check{\mathbf{h}}_{v} = \mathbf{W}^{'}_v \mathbf{v}_t \in \mathbb{R}^{V \times H}.
\end{align}
where $\check{\mathbf{h}}_{v}$ represents the embeddings for all vessel classes, rather than just the selected vessel class as used in Eq.~\ref{eq:encoder_vessel_embed} for the encoder-only approach. Note the $\mathbf{W}^{'}_v \in \mathbb{R}^{H\times D_v}$ is distinct from $\mathbf{W}_v$. Next, we define the overall embedding for the decoder:
\begin{align}
    \mathbf{h}_{\text{embed}} = 
        \begin{bmatrix}
        \check{\mathbf{h}}_p \\
        \check{\mathbf{h}}_v \\
        \mathbf{h}_{\text{BOS}}
        \end{bmatrix} \in \mathbb{R}^{\bar{N}\times H},
\end{align}
where $\mathbf{h}_{\text{BOS}} \in \mathbb{R}^{H}$ represents the embedding for the ``beginning of service'' (BOS). This vector is randomly initialized and remains static throughout the entire service generation process. The notation $\bar{N} = P+V+1$ reflects the total dimensionality of the embedding, where $P$ is the number of ports, $V$ is the number of vessel classes, and the additional 1 corresponds to the BOS. It’s important to note that the embeddings $\check{\mathbf{h}}_p$ and $\check{\mathbf{h}}_p$ vary with each step $t$, but remain constant across all sub-step $\tau$'s. For simplicity, we have omitted the $t$ subscripts in this equation.

\begin{figure}[htp]
    \centering
    \includegraphics[width=1\textwidth]{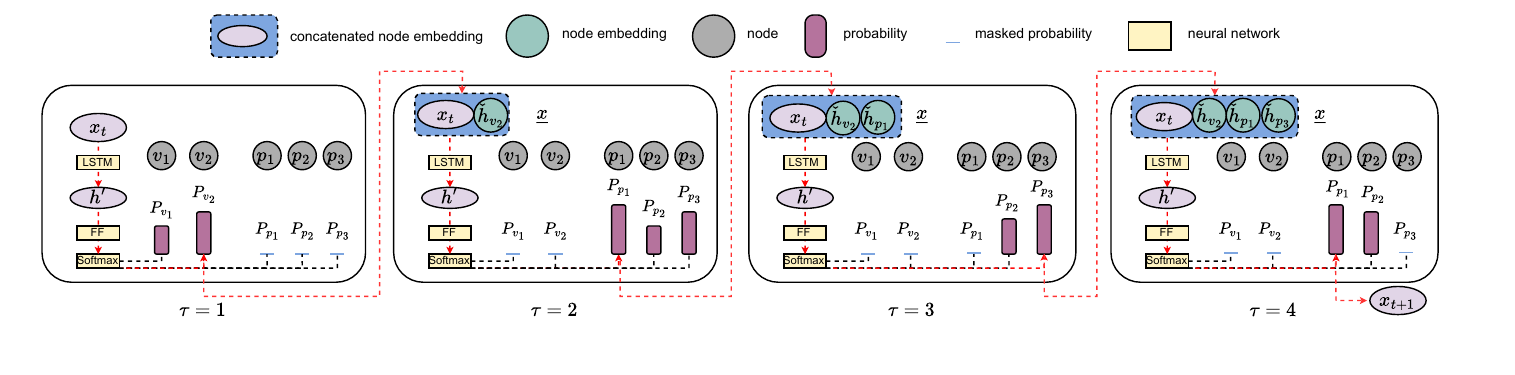}
    \caption{LSTM-based decoder for NDP. The decoder takes in $\mathbf{x}_{t}$ generated from the previous steps and builds $\underline{\mathbf{x}}$ sequentially. The example shows how a full service $A_t = (v_2,p_1,p_3)$ is generated sequentially over $\tau=1,2,3,4$ within step $t$. Note that at $\tau=4$, $p_1$ is selected again, which closes the circle and ends $A_t$.}
    \label{fig:decoder}
\end{figure}

Figure~\ref{fig:decoder} illustrates how the agent progresses through each sub-step $\tau$ within a step $t$, using Long Short-Term Memory (LSTM) to guide the rollout process. This can be mathematically expressed as follows:
\begin{align}
    \mathbf{h}^{\prime}_{1}, \mathbf{h}_{1}, \mathbf{c}_{1} &= \textbf{LSTM}^{(1)}\left(\mathbf{x}_{t}, \mathbf{h}_{0}, \mathbf{c}_{0}\right), \label{eq:lstm_1}\\
    \mathbf{h}^{\prime}_{\tau}, \mathbf{h}_{\tau}, \mathbf{c}_{\tau} &= \textbf{LSTM}^{(\tau)}\left(\underline{\mathbf{x}}, \mathbf{h}_{\tau-1}, \mathbf{c}_{\tau-1}\right). \label{eq:lstm_tau}
\end{align}
%
Here, $\mathbf{x}_t \in \mathbb{R}^{H \times n_0}$ represents the embeddings of all vessels and ports selected in prior services up to step $t$, while $\underline{\mathbf{x}}\in \mathbb{R}^{H \times n}$ extends this by including the embeddings of vessels and ports selected up to sub-step $\tau$, where $n = n_0 + \tau$. The cell state of the LSTM at sub-step $\tau$, denoted by $\mathbf{c}_{\tau}$, is initialized as
\begin{align}
    \mathbf{c}_{0} = \mathbf{h}_{p, \text{global}}^{(L)} \in \mathbb{R}^{H},
\end{align}
where $\mathbf{h}_{p, \text{global}}^{(L)}$ is the global embedding defined in Eq,~\ref{eq:encoder_embedding}. The hidden state of the LSTM at sub-step $\tau$, represented by
$\mathbf{h}_{\tau}$
, is initialized as:
\begin{align}
    \mathbf{h}_{0} = \frac{1}{P} \sum_{i=1}^{P} (\check{\mathbf{h}}_{p})_{i} \in \mathbb{R}^{H}.
\end{align}

Once the output of the LSTM $\mathbf{h}^{\prime}_{\tau} \in \mathbb{R}^{H\times n}$ is generated through Eq.~\ref{eq:lstm_tau}, it is passed through a fully connected feed-forward (\textbf{FF}) layer with ReLU activation, which transforms the embedding from dimension $H$ to $\bar{N}$. To enhance stability during training, layer normalization (\textbf{LN}) is applied to the resulting embeddings:
\begin{align}
    \widehat{\mathbf{h}}^{\tau} &= \textbf{LN}\left(\textbf{FF}(\mathbf{h}^{\prime}_{\tau})\right), \\
    \widehat{\mathbf{h}}^{\tau} &= [\widehat{\mathbf{h}}^{\tau}_{1}, \dots, \widehat{\mathbf{h}}^{\tau}_{n}] \in \mathbb{R}^{\bar{N} \times n}.
\end{align}
The last embedding vector $\widehat{\mathbf{h}}^{\tau}_{n} \in \mathbb{R}^{\bar{N}}$ is then processed through a Softmax layer to produce the final output probabilities:
\begin{align}
    \widetilde{\mathbf{h}}_{\tau} &= \frac{e^{\widehat{\mathbf{h}}^{\tau}_{n}}}{\sum_{j=1}^{\bar{N}} e^{(\widehat{\mathbf{h}}^{\tau}_{n})_{j}}} \in \mathbb{R}^{\bar{N}}.
\end{align}

The probability distribution is filtered through a masking rule before it is used to sample the index of the vector which represents the next vessel or port in the service. The embedding of the selected vessel or port is then appended to $\underline{\mathbf{x}}$, which will serve as the input vector for the subsequent sub-step:
\begin{align}
    i &\sim \mathbb{P}\left(\textbf{mask}(\widetilde{\mathbf{h}}_{\tau})\right), \\
    \underline{\mathbf{x}} & \leftarrow [\underline{\mathbf{x}}, (\mathbf{h}_{\text{embed}})_{i}].
\end{align}
Here, $i$ represents the index of vector $\widetilde{\mathbf{h}}_{\tau}$. The masking rule operates as follows: during the very first sub-step within step $t$, all ports are masked to allow for vessel selection. From the second sub-step onward, the ports are unmasked while the vessels are masked. Ports that have already been visited in step $t$ remain masked, except for the first port, as revisiting it indicates the completion of a service generation. Noted that the ``begin of service'' (BOS) embedding is only unmasked at the first sub-step ($\tau=1$) of the initial step ($t=1$).

Importantly, both $\mathbf{x}_t$ and $\underline{\mathbf{x}}$ represent the same set of embeddings, capturing the vessels and ports selected at various stages. When carried over across different steps, it is referred to as $\mathbf{x}_t$, whereas when carried over across sub-steps within the same step, it is denoted as $\underline{\mathbf{x}}$. 

The sub-steps in each action are generated autoregressively, with our neural network architecture applying the chain rule to factorize the probability of generating a service at step $t$ as:
\begin{align}
    \pi_{\theta}(A_t|S_t)=\prod_{\tau=1}^{n_{\tau}}\mathbb{P}(A_t(\tau)|A_t(\tau^{\prime}<\tau),S_t),
\end{align}
where $n_{\tau}$ represents the total number of sub-steps within step $t$. The term $A_t(\tau)$ refers to the selection made at sub-step $\tau$, while $A_t(\tau^{\prime}<\tau)$ denotes all selections made in the previous sub-steps leading up to $\tau$.

\section{Policy Optimization}\label{sec:optimization}
To optimize our policy network $\pi_\theta$, we employ policy gradient methods, which iteratively refine the policy to maximize the reward. Policy gradient methods form a broad class of reinforcement learning algorithms that directly improve the policy $\pi_\theta$ by rewarding actions that lead to higher-value outcomes based on sampled trajectories. In this work, we utilize an enhanced variant of the standard policy gradient algorithm known as Proximal Policy Optimization (PPO) \cite{schulman2017proximal}. PPO introduces a clipped surrogate objective that significantly enhances the stability of the learning process. Detailed hyperparameter settings for PPO are provided in Appendix~\ref{app:hyperparam}.

\section{Experiments}\label{sec:result}

We conduct experiments to evaluate the performance of the proposed RL approach on the LINERLIB benchmark, demonstrating that the RL-based solution for the NDP is a promising alternative to MIP and heuristic methods. Notably, Google ORTools\footnote{https://developers.google.com/optimization/service/shipping/benchmarks/lsndsp} recently published their own benchmark results on the LINERLIB dataset, which we include in our comparison where applicable.

To focus on benchmarking the solution quality for the NDP, we use the MCF algorithm introduced in Appendix~\ref{app:heuristic_mcf} to evaluate all network solutions, including those generated by our RL-based method, LINERLIB, and ORTools. For a consistent comparison, we decompose all multi-loop services, such as butterfly services, into simple services. Additionally, we relax the hard limit on the number of vessels, treating it as a soft constraint with an associated penalty, to further facilitate the comparison.

Unless otherwise stated, all experiments, including both training and inference, are performed on an A100 GPU. As described in Section~\ref{sec:optimization}, we employ the standard PPO algorithm for policy optimization during training. For detailed hyper-parameter settings, please refer to Appendix~\ref{app:hyperparam}.


\subsection{Result on Baltic Instance}\label{subsec:baltic}
In this section, we evaluate the performance of our RL-based NDP approach on the Baltic instance from the LINERLIB dataset, which consists of 12 ports (i.e., vertices in the graph). Training and validation are conducted on separate datasets with a total of 16,000 instances, where the demand quantities were perturbed from the original LINERLIB Baltic instance (with a factor of $\pm 10\%$), while the origins and destinations of the demand remained unchanged. For a detailed description of the perturbation process, please refer to Appendix~\ref{app:perturb}. The test set consists of a single data point—the actual LINERLIB Baltic instance—to ensure a fair comparison with publicly available benchmark solutions. It is worth noting that we primarily report results from the encoder-decoder RL-based solution, as the encoder-only variant produced nearly identical results in this case.

Table~\ref{table:breakdown} provides a detailed profit breakdown for the RL-based solution and compares it with the LINERLIB benchmark. Notably, if a solution utilizes fewer vessels than the available fleet, a profit is gained based on the time charter rate. Conversely, if the solution requires more vessels than available, additional costs are incurred.

\begin{table}[htp]
\caption{Profit (\$) breakdown of the RL-based NDP solution and LINERLIB solution.}
\centering
\begin{tabular}{l S[table-format=7.0] S[table-format=7.0]} 

                                & {RL-based Solution} & {LINERLIB Solution}  \\ \toprule \toprule
Revenue                         & 3688028         & 3687260  \\ \hline
Unused vessel profit            & -12596          & 6823     \\ \hline
Vessel used                     & ~~~~~~~~~{6.34}      & ~~~~~~~~~{5.72}      \\ \hline
Vessel service cost             & 267898          & 245176   \\ \hline
Voyage cost and fee             & 634729          & 689083   \\ \hline
Handling and transshipment cost & 2116377         & 2109876  \\ \hline
Rejected demand penalty         & 380000          & 389000   \\ \hline
\textbf{Total net profit}       &~~~\textbf{276,428} &~~~\textbf{260,948} \\ \bottomrule \bottomrule
\end{tabular}
\label{table:breakdown}
\end{table}

The RL-based approach utilizes 2.03 Feeder 800 vessels (with a capacity of 800 FFEs) and 4.31 Feeder 450 vessels (with a capacity of 450 FFEs), while the LINERLIB solution employs 2.14 Feeder 800 vessels and 3.58 Feeder 450 vessels. Any over- or under-utilization of vessels is accounted for in the ``unused vessel profit,'' where the LINERLIB solution shows a positive profit, while the RL-based solution incurs a loss. However, the RL-based solution is able to satisfy more demand, resulting in a smaller penalty for rejected demand due to the higher vessel utilization. Despite these nuances, the RL-based solution ultimately achieves a higher net profit compared to the LINERLIB solution. For a visual comparison of the network designs, refer to Figures~\ref{fig:design1} and \ref{fig:design2} in Appendix~\ref{app:baltic}, which show the networks produced by the RL-based and LINERLIB solutions, respectively.

\subsection{Experiments on Other Instances}\label{subsec:optimizer}
In this section, we explore the potential of using the RL-based NDP solution as an optimizer, where the training process of the RL agent functions as a traditional optimization solver. In this scenario, there is no distinction between the training and inference phases.

We extend the experiments to include two additional instances from LINERLIB: West Africa (WAF) and World Small. These instances contain 20 and 47 ports, respectively. A performance comparison between the two RL-based NDP approaches—encoder-only and encoder-decoder—and the benchmark solutions is presented in Table~\ref{table:performance}. Both LINERLIB and ORTools solutions are listed as benchmarks, with ORTools results being available only for the World Small instance. Please refer to Appendix~\ref{app:waf_ws} for a visual comparison of the networks designed by the RL-based solution and the LINERLIB solution. 

To ensure a fair comparison, both the LINERLIB and ORTools solutions are evaluated using our MCF algorithm introduced in Appendix~\ref{app:heuristic_mcf}.

\begin{table}[htp]
\centering
\caption{Max profit (in million \$) of RL-based NDP approaches on LINERLIB problem instances.}
\begin{tabular}{lSSS}

                    & {Baltic (n=12)} & {WAF (n=20)}   & {World Small (n=47)}  \\ \toprule \toprule
Encoder-only (RL)    & 0.28   &  ~~\textbf{5.60} & \textbf{42.73} \\ \hline
Encoder-decoder (RL) & ~~\textbf{0.29}   & ~~\textbf{5.60}  & 42.32  \\ \hline
LINERLIB             & 0.26   & 5.20  & 32.28  \\ \hline
ORTools              & ~~{N/A}  & ~~{N/A}  & 40.10  \\ \bottomrule \bottomrule
\end{tabular}
\label{table:performance}
\end{table}

Both RL-based NDP approaches yield higher profits compared to the benchmark solutions, demonstrating the clear potential of using the RL-based solutions as a viable alternative solver for the Network Design Problem.

\subsection{Solve Time Comparison}\label{subsec:solvetime}
Table~\ref{table:time} reports both the inference time and training time for the RL-based NDP solution on the LINERLIB instances we experimented on. The solve time is benchmarked with the LINERLIB solution on corresponding instances. Note that for the RL-based NDP solution, the inferences are all run on an Apple M2 CPU with 12 cores, while the trainings are conducted on an A100 GPU. We only report the encoder-only approach as the RL-based NDP solution in Table~\ref{table:time} given that the encoder-decoder approach yields an inference time similar to that of the encoder-only approach. 

\begin{table}[htp]
\centering
\caption{Inference time (in seconds) of the proposed encoder-only approach on different LINERLIB instances.}
\begin{tabular}{l
                S[table-number-alignment = right, table-figures-integer = 3, table-figures-decimal = 2] 
                S[table-number-alignment = right] 
                S[table-number-alignment = right] 
               }

                                          & {Baltic (n=12)} & {WAF (n=20)}   & {World Small (n=47)} \\ \toprule
\toprule
RL-based NDP inference                    & 0.03            & 0.11           & 0.60                 \\ \hline
Environment simulation + RL-based inference & ~~\textbf{0.22}   & ~~\textbf{0.40}  & \textbf{15.04}       \\ \hline
LINERLIB benchmark                        & ~~~{300}             & ~~~{900}            & \hspace{-0.1cm}{10,800}                \\ \hline
RL training time (excluding inferences)   & ~~~{720}             & ~{4000}           & \hspace{-0.3cm}{360,000}               \\ 
\bottomrule
\bottomrule
\end{tabular}
\label{table:time}
\end{table}

It is important to clarify that the ``RL-based NDP inference'' time in Table~\ref{table:time} includes both the generation of the full set of services and the execution of the underlying MCF algorithm, whereas the ``environment simulation + RL-based inference'' time also accounts for the setup and updates of the environment. When comparing the inference time of the RL-based solution to the solve time of the LINERLIB solution, which uses a MIP solver, we observe an approximate 1000x speedup across the three problem instances tested.

\subsection{Solution Robustness against Variations in the Problem Instance}\label{subsec:robust}
In real-world applications beyond the academic scope of the LSNDP, schedulers often face disturbances on short notice that can have long-term impacts. Examples include trade wars, which affect demand quantities, or pirate activities, which influence the availability of certain ports or routes in network design. Consequently, having an algorithmic tool that can quickly generate optimal network designs for a variety of perturbed problem instances is of immense value.

In this part of the experiment, we evaluate the effectiveness of our RL-based NDP solution in handling a large set of problem instances perturbed from a common baseline. Additionally, we explore how enhancing the RL agent by exposing it to these perturbed instances during training improves its performance compared to the baseline RL agent, which is trained on a single problem instance (as used in Section~\ref{subsec:optimizer}). For the enhanced RL agent, training and validation are conducted on separate datasets totaling 80,000 instances, where demand quantities are perturbed by $\pm 10\%$ from the original LINERLIB Baltic instance (matching the perturbation level in Section~\ref{subsec:baltic}). During inference, 100 different test instances are randomly generated, and for each instance, the RL agent produces 100 network designs. From these designs, the maximum profit (i.e., reward) is selected for each instance. The mean and standard deviation of these maximum profits across the 100 test instances are then reported.

To further examine the agent’s robustness, we increase the perturbation level in both the training and test datasets from 10\% to 50\%, assessing how much additional improvement the enhanced RL agent offers over the baseline. Notably, the perturbation levels are kept consistent between training and testing; for instance, when evaluating performance on a dataset with 50\% perturbation, the RL agent is trained on data with the same 50\% perturbation.

The first row of Table~\ref{table:perturb} compares the mean profit value across 100 test instances with a 10\% perturbation between the enhanced and baseline RL agents. On average, the enhanced RL agent trained on perturbed instances achieves a \$257,945 higher profit than the baseline RL agent trained on a single instance. The second row shows that with the demand quantity perturbation increased to 50\%, both agents perform worse, but the profit uplift from the enhanced RL agent rises to \$401,005. It is important to note that these results are based on the encoder-decoder RL approach, though we expect similar trends for the encoder-only approach.

\begin{table}[htp]
\caption{Mean and standard deviation (in parenthesis) of profits in \$ over 100 test instances for enhanced RL agent and baseline RL agent at corresponding perturbation levels.}
\centering
\begin{tabular}{lSS}
                       & {\makecell{Enhanced agent \\ trained on perturbed instances}} & {\makecell{Baseline agent \\trained on single instance}} \\ \toprule \toprule
Test dataset (10\% perturbation) & 
{274,387.17   ~~~  (5,835.62)}     & ~~~{16,441.79 (25,430.68)}      \\ \hline
Test dataset (50\% perturbation) & ~~{78,215.53 ~~(34,042.51)}    & {-322,789.09 (56,312.07)}      \\ \bottomrule \bottomrule
\end{tabular}%
\label{table:perturb}
\end{table}

\subsection{Discussion}\label{subsec:discussion}
The experiments conducted in this section demonstrate the effectiveness of our RL-based NDP solution in two significant ways. Firstly, when evaluated on the Baltic instance from the LINERLIB dataset, the RL-based solution generates near-optimal results and compares favorably against the benchmark solutions. Secondly, when applied as an optimizer on previously unseen instances, such as the Baltic, West Africa (WAF), and World Small datasets, the RL-based approach continues to deliver competitive performance without the need for retraining. This motivates the use of reinforcement learning based methods to learn general, competitive policies that can potentially deliver high-quality solutions on new instances. 

However, a few limitations should be noted, stemming from both the computational resources and the experimental setup. The most significant limitation arises from the underlying heuristic multi-commodity flow (MCF) algorithm, which serves as a key part of the reward function evaluator for the RL agent. Unfortunately, we do not have access to the state-of-the-art MCF implementations used by the benchmarks. This discrepancy between our MCF implementation and those used in the benchmarks means the associated NDPs are effectively different problems. As a result, we evaluate all benchmark solutions using our MCF implementation.

Additionally, it's important to note that the NDP definition has been relaxed to better align with our RL-based approach. For instance, we consider only simple services in the network design and relaxed the hard limit on the number of vessels to a soft constraint with penalties. These modifications explain why the LINERLIB and ORTools solutions reported here may differ from those found in other literature. According to industry experts, these relaxations have only a limited impact on the solution quality. Nonetheless, tightening these constraints and preparing the solution for an end-to-end benchmark on the full LSNDP would be a valuable next step.

Our current experiments cover three out of the seven instances in the LINERLIB dataset. Expanding the experiments to include all instances, particularly the World Large instance (the largest in the dataset), would further test the scalability of the approach. Moreover, the perturbations in this study are limited to demand quantities. Extending the perturbations to include the origin and destination of the demand, the number of available vessels in each class, and the inclusion of specific ports would provide a more comprehensive demonstration of the solution's generalizability.

\section{Conclusion and Future Work}\label{sec:conclusion}
In this paper, we propose a model-free RL-based framework to address the network design aspect of the Liner Shipping Network Design Problem (LSNDP). By leveraging a heuristic-based multi-commodity flow (MCF) solver as part of the evaluator function, our approach can solve the LSNDP in an end-to-end fashion. This work marks the first attempt to approach LSNDP through a method distinct from traditional operations research (OR) techniques. Our framework demonstrates scalability with problem size and achieves competitive results on the LINERLIB benchmark. We have shown that our approach offers value in two key ways: it can rapidly generate near-optimal solutions for problem instances perturbed from the training data or be utilized as an optimizer, delivering effective performance on unseen problem instances without requiring prior training.

Our approach introduces a novel paradigm for solving LSNDP compared to conventional OR methods, which typically require equal computational effort for each new problem instance. In contrast, our method front-loads the computational work during the training phase, while enabling rapid inference to new instances. This makes the solution ideal for problems like LSNDP, where long-term plans are frequently disrupted by unexpected events or frequent data updates. In terms of real world applications, this enables rapid ``tactical'' changes by reacting to real world dynamics.

Looking ahead, there are several opportunities to enhance the encoder-decoder architecture. Replacing the LSTM with a transformer-based architecture could allow the network to handle larger and more complex use cases. Additionally, as mentioned in Section~\ref{subsec:solvetime}, the MCF algorithm accounts for a significant portion of the runtime during both training and inference. A faster MCF implementation would further improve training efficiency and reduce thhe training wall clock time. Exploring a Graph Neural Network (GNN)-based surrogate for MCF is another promising avenue to speed up reward function evaluation.

In terms of training strategies, there are several paths to explore. One interesting direction is the application of reward shaping, as discussed by \cite{ng1999policy}, to enhance training efficiency. Reward shaping provides additional signals to guide the RL agent toward an optimal policy, especially when only terminal rewards are available during policy exploration. Introducing penalties in the reward function could also help guide the agent’s behavior; for instance, adding a service length penalty could encourage the agent to select shorter routes that optimize transshipment usage.

Another area worth exploring is the inherent symmetry of the LSNDP, where the reward remains unchanged if ports are rotated within a service. Inspired by OR techniques, which often limit symmetry in the search space to improve solving speed, symmetry can also be leveraged in neural combinatorial optimization, as demonstrated by \cite{kwon2020pomo} and \cite{kim2022sym}. Adapting these techniques to our RL framework could improve sample efficiency for LSNDP.

Other reinforcement learning algorithms that favor exploration and improve sample efficiency could potentially improve performance over PPO. For example, Soft Actor-Critic (SAC, \cite{haarnoja2018soft}), an off-policy actor-critic algorithm, maximizes both expected reward and entropy, promoting more effective exploration. Additionally, Monte Carlo Tree Search (MCTS, \cite{kocsis2006bandit}) based methods when combined with neural networks \cite{AlphaZero} have proved to be effective model-based approaches and have yielded superhuman performance in deterministic environments. RL algorithms designed to scale with problem size (\cite{drori2020learning}) and generalize across a variety of instances (\cite{fu2021generalize}) may also align with broader business needs beyond LSNDP. Adapting these techniques to train models on smaller instances and transfer the learned policy to larger problems would be a valuable extension of this work.

\section*{Acknowledgement}

We want to extend our heartfelt gratitude to Carl Mikkelsen, Jimmy Paillet, and Torkil Kollsker for their guidance in designing the multi-commodity flow heuristics used in this paper. Our deep appreciation also goes to Josh Zhang for his support in enhancing the algorithm's efficiency. We are especially grateful to Sina Pakazad for generously providing the necessary computing resources. We acknowledge the insightful contributions from Henrik Ohlsson, Nikhil Krishnan, Sravan Jayanthi, and Fabian Rigterink during the early stages of problem formulation. Additionally, we are thankful for the project management support from Marius Golombeck, Abhay Soorya, and Mehdi Maasoumy. We would like to express our gratitude to C3.ai for their funding and platform infrastructure. If you are interested in providing funding or collaborating on future research in this area, we encourage you to reach out to C3.ai.

\clearpage

\bibliographystyle{abbrvnat}
\bibliography{references}

\clearpage

\appendix
\section{LSNDP details}\label{app:lsndp}
\subsection{Dataset}\label{subsec:linerlib}
\cite{brouer2014base} offers a comprehensive introduction to the LSNDP benchmark suite, LINERLIB. Here, we provide a brief overview to establish the context for the problem we aim to address using reinforcement learning. At a high level, a liner shipping network comprises a fleet of vessels $V$ deployed across rotations or services $S$ to satisfy a set of commodity demands $D$, normalized to a weekly frequency. Visualizing the network as a graph, ports can be seen as vertices, with edges $E$ representing the connections between them. Each service $s \in S$ involves a rotation through a sequence of ports $s_P$, is assigned a specific subset of vessels $s_V$, and includes a set of legs $s_E$ that define the route. Below, we provide a more detailed breakdown of each of these elements.

LINERLIB includes a predefined set of ports $P$ that vessels can access. Each port $p$ within this set is characterized by the following features:
\vspace{-0.4cm}
\begin{table}[!htb]
\begin{tabu} to \linewidth {lX}
\multicolumn{2}{l}{} \\
$p$~~~~~~~~~~~~~ & Port ID, represented by UNLOCODE.\\
$p_{\text{f}}$~~~~ & Fixed cost per port call, the cost in USD for each vessel call at this port. \\
$p_{\text{v}}$~~~~ & Variable cost per port call, the additional cost in USD per FFE for visiting this port, based on the vessel's capacity.\\
$p_{\text{t}}$~~~~ & Transshipment cost per FFE, the cost in USD per FFE for transferring cargo across different services at this port. \\
\end{tabu}
\end{table}
\vspace{-0.2cm}

A fleet of vessels $V$ contains different vessel classes $v^F$, each with different capacities and characteristics. Each vessel class has a finite number of vessels available for deployment. A vessel $v \in v^F$ is characterized by the following features:
\vspace{-0.4cm}
\begin{table}[!htb]
\begin{tabu} to \linewidth {lX}
\multicolumn{2}{l}{} \\
$v_{\text{cap}}$~~~~ & Capacity, the maximum number of FFEs the vessel can carry at once.\\
$v_{\text{n}}$~~~~ & Quantity, the total number of available vessels of class $v$.\\
$v_{\text{TC}}$~~~~ & TC rate, the daily cost of renting or operating the vessel.\\
$v_{\text{s}}$~~~~ & Design speed, the vessel's standard sailing speed. \\
$v_{\text{fs}}$~~~~ & Fuel consumption at design speed, the vessel's daily fuel consumption (converted to \$) when sailing at design speed. \\
$v_{\text{fi}}$~~~~ & Fuel consumption while idling, the vessel's daily fuel consumption (converted to \$) when idle at the port. \\
$v_{\text{Suez}}$~~~~ & Suez fee, the fee for passing through the Suez Canal. \\
$v_{\text{Panama}}$~~~~ & Panama fee, the fee for passing through the Panama Canal. \\
\end{tabu}
\end{table}
\vspace{-0.2cm}

Each port in $p$ also includes data on latitude and longitude coordinates (which we omitted earlier for brevity). The LINERLIB dataset provides distance information, including whether the route passes through the Panama or Suez canals. This distance data corresponds to the edge $e \in E$ in the graph and is described by the following features:
\vspace{-0.4cm}
\begin{table}[!htb]
\begin{tabu} to \linewidth {lX}
\multicolumn{2}{l}{} \\
$e_{\text{o}}$~~~~ & Origin port, the Port ID in UNLOCODE.\\
$e_{\text{d}}$~~~~ & Destination port, the Port ID in UNLOCODE.\\
$e_{\text{dist}}$~~~~ & Distance, the distance between the origin port and the destination port, measured in nautical miles.\\
$e_{\text{Suez}}$~~~~ & Suez traversal, a flag indicating whether the sailing route passes through the Suez Canal. A value of 1 signifies the route uses the Suez Canal; 0 otherwise. \\
$e_{\text{Panama}}$~~~~ & Panama traversal, a flag indicating whether the sailing route passes through the Panama Canal. A value of 1 signifies the route uses the Panama Canal; 0 otherwise. \\
\end{tabu}
\end{table}
\vspace{-0.2cm}

At last, each commodity demand $d \in D$ is characterized by the following features:
\vspace{-0.4cm}
\begin{table}[!htb]
\begin{tabu} to \linewidth {lX}
\multicolumn{2}{l}{} \\
$d_o$~~~~~~~~~~~ & Origin port, the Port ID in UNLOCODE.\\
$d_d$~~~~ & Destination port, the Port ID in UNLOCODE.\\
$d_R$~~~~ & Revenue, generated per unit FFE transported.\\
$d_q$~~~~ & Quantity, demand quantity in FFE per week. \\
$Y_{d}$~~~~ & Penalty if rejected, penalty for rejection of this demand, which is set to \$1000.\\
\end{tabu}
\end{table}
\vspace{-0.2cm}

Note that we have only listed the dataset elements relevant for solving the LSNDP with transshipment, rejected demand, and fractional vessel assignments. For a complete description of the dataset, please refer to \cite{brouer2014base}. Throughout this paper, when we refer to an ``instance'', we mean a specific LSNDP setup with a defined set of commodities, available ports, edges, and fleet, which collectively determine the characteristics of the shipping network.

\subsection{Multi Commodity Flow}\label{subsec:mcf}
The maximum profit Multi-Commodity Flow Problem seeks to determine the optimal flow of multiple commodities through a capacitated network to maximize total profit. Each commodity has a specific origin and destination and moves through the network's edges, constrained by capacity limits. Additionally, each unit of flow for a commodity may generate a defined revenue. The goal is to allocate flows for all commodities in a way that maximizes the overall profit while adhering to the network's capacity constraints. The capacity limits of the network are defined by the designed rotations or services from the associated network design problem, which will be discussed later. The capacity constraints on each edge of the network are derived from the capacities of the vessels assigned to those routes.

Here, we omit the specific details of the constraints and focus only on the objective function. For a comprehensive description of the complete problem formulation, please refer to \cite{brouer2014base}.

\begin{align}
    \text{Maximize} \quad \eta &= R_{\text{total}} - C_{\text{reject}} -  C_{\text{handle}} - C_{\text{NDP}}, \label{eq:mcf_profit}
\end{align}
where $\eta$ is the profit, $R_{\text{total}}$ represents the total revenue generated, $C_{\text{reject}}$ represents the penalty associated with rejected demand, and $C_{\text{handle}}$ denotes the handling costs. Detailed descriptions of these terms are provided in the equations below. Note that $C_{\text{NDP}}$ is the fixed cost of establishing all services in the network, independent of the decisions made within the multi-commodity flow problem. This fixed cost is determined by the network design and will be discussed in detail later.
\begin{align}
    R_{\text{total}} & = \sum_{d \in D} d_R \left( \sum_{\forall e | e_d = d_d} f_e^d\right), \label{eq:total_revenue}\\
    C_{\text{reject}} & = Y_d\sum_{d \in D}\left( d_q -\sum_{\forall e | e_d = d_d} f_e^d\right), \label{eq:rejected_demand}\\
    C_{\text{handle}} & = \sum_{p \in P} p_{\text{l}} \left( \sum_{\forall e | e_d = d_d= p}f_e^d + \sum_{\forall e | e_o = d_o = p}f_e^d \right) 
    + \sum_{p \in P}p_{\text{t}} \sum_{\substack{\forall e', e'' \in S_E | \\ e'_d=p, e''_o=p, e'_d\neq d_d}} \left( f_{e'}^{d} - f_{e''}^{d}\right). \label{eq:handling_cost}
\end{align}
Here, $f_e^d$ is a decision variable within MCF, which represents the quantity of commodity demand $d$ that flows through edge $e$, while $e'$ and $e''$ refer to two edges within the same service. The remaining notation is detailed in Appendix~\ref{subsec:linerlib}. It is important to note that the handling cost, as described in Eq.~\ref{eq:handling_cost}, has two components: the first is the cost associated with onloading and offloading, and the second is the transshipment cost. It is important to note that, in contrast to the fixed cost $C_{\text{NDP}}$, the terms in Eqs.~\ref{eq:total_revenue}, \ref{eq:rejected_demand}, and \ref{eq:handling_cost} represent variable costs, with revenue broadly considered as a form of negative cost.

The MCF problem is known to be NP-hard (see Theorem 6.2 in \cite{brouer2014base}). Rather than solving it using a MIP formulation, we employ a fast, greedy heuristic-based approach, which is detailed in Appendix~\ref{app:heuristic_mcf}.

\subsection{Network Design Problem}\label{subsec:ndp}
The network design problem (NDP) focuses on identifying the optimal set of services for a given LSNDP instance to maximize the overall profitability of the shipping network. However, the ultimate profitability of the network is determined by the Multi-Commodity Flow (MCF) solution, as discussed previously. The primary objective of the NDP is to develop a network design that defines the capacities on the edges of the services, which are used as constraints in MCF. The fixed cost associated with the designed network corresponds to the $C_{\text{NDP}}$ term in Eq.~\ref{eq:mcf_profit} and is calculated as follows:
\begin{align}
    C_{\text{NDP}} = C_{\text{service}} + C_{\text{unused}} + C_{\text{voyage}},
\end{align}
where $C_{\text{service}}$ represents the vessel service cost, accounting for the total cost of renting or operating all vessels assigned to the services. $C_{\text{unused}}$ captures the cost (or profit) of unused vessels. If the generated services do not utilize all available vessels, the remaining vessels can be rented out at the time charter (or TC) rates. Conversely, if the services require more vessels than are available, additional vessels must be acquired at the same rate. The term 
$C_{\text{voyage}}$ refers to the voyage cost, which includes fuel costs, port calling costs, and canal fees associated with operating the vessels to support the services. The detailed breakdown of these terms is provided in the equations below:
\begin{align}
    C_{\text{service}} & = \sum_{s \in S} \sum_{v \in s_V} n_{v,s}\cdot v_{\text{TC}}, \label{eq:service_cost}\\
    C_{\text{unused}} & = - \sum_{v \in V} \left( v_n - \sum_{r \in R}n_{v,r}\cdot v_{\text{TC}}\right),\label{eq:unused_vessel_cost}\\
    C_{\text{voyage}} & = \sum_{s \in S} \sum_{p \in s_{P}} \sum_{v \in s_V} \left( p_{\text{f}} + p_{\text{v}}\cdot v_{\text{cap}} \right)\cdot n_{v,s} + \sum_{s \in S} \sum_{v \in s_V} \left( \frac{\sum_{e\in s_E}e_{\text{dist}}}{v_s}\cdot v_{\text{fs}} + \sum_{p \in s_P}1\cdot v_{\text{fi}}\right)\cdot n_{v,s} \nonumber\\
    & + \sum_{s \in S} \sum_{v\in s_V} \sum_{e\in s_E}\left( e_{\text{suez}}\cdot v_{\text{suez}} + e_{\text{panama}} \cdot v_{\text{panama}}\right). \label{eq:voyage_cost}
\end{align}
Here, $C_{\text{voyage}}$ accounts for various voyage-related expenses, including fixed and variable port fees, fuel consumption during sailing, and costs associated with passing through the Suez and Panama canals. The variable $n_{v,s}$ represents the number of vessels in a given class that are assigned to service $s$, making it a key decision variable in the NDP. The remaining notation is detailed in Appendix~\ref{subsec:linerlib}. Since the problem setup assumes that vessels operate strictly at their designed speed and allows for fractional vessel assignments, the required number of vessels can be calculated as a function of the total distance of the service and the vessel's designed speed. As a result, there is no expected idle time for the vessels except for the required 1 day at each port.

To integrate the NDP and MCF, we first generate a network schedule through the NDP, which is subsequently used as input for the MCF to calculate the associated revenues and penalties. The combined objective value thus includes both the static network costs derived from the NDP and the variable flow costs calculated from the MCF. This combined objective serves as the reward function, which is used to evaluate the performance of our complete algorithm.

\section{Heuristic MCF Details}\label{app:heuristic_mcf}
As outlined in Appendix~\ref{subsec:mcf}, the Multi-Commodity Flow (MCF) problem seeks to identify optimal cargo routes that maximize profit within a capacitated network. Instead of solving the MCF to optimality using a Mixed-Integer Programming (MIP) approach, we propose a faster, heuristic-based method that employs a greedy sequential commodity flow strategy, building on a graph representation of the liner shipping network.

\begin{figure}[htp]
    \centering
    \includegraphics[width=0.9\linewidth]{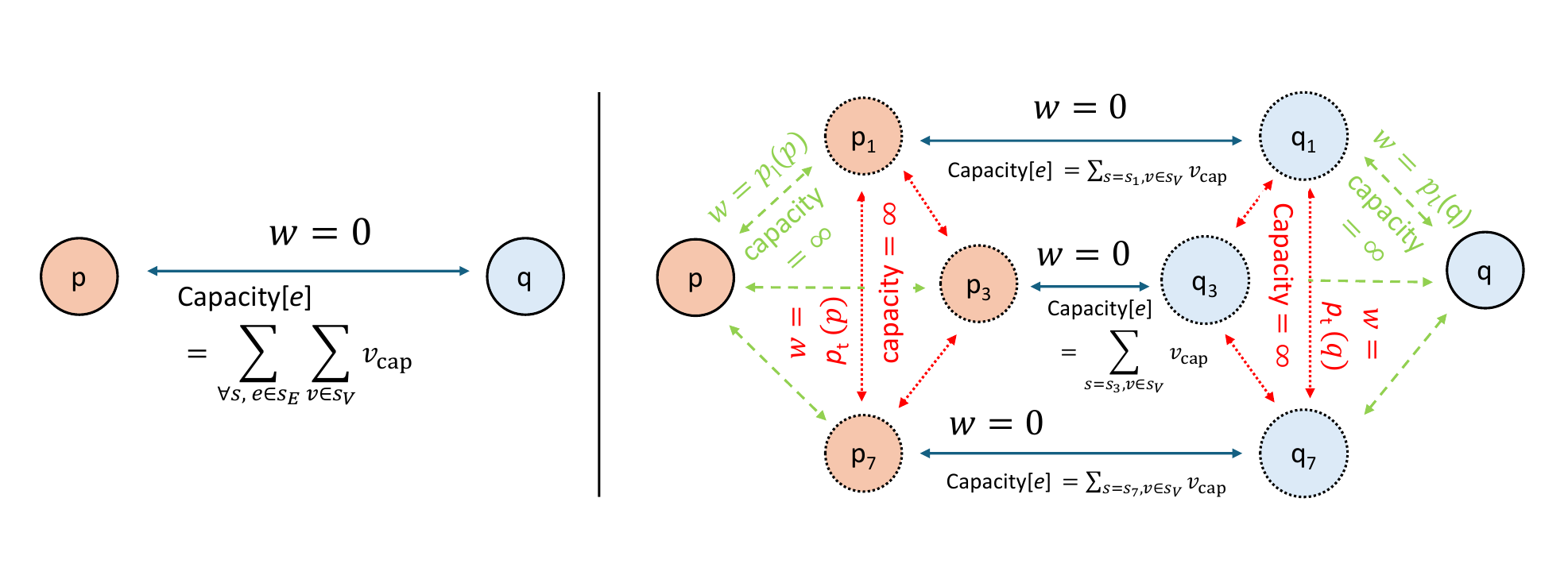}
    \caption{Expanded graph representation for an edge connecting ports $p$ and $q$. The expanded representation on the right side of the figure with proxy nodes fully captures all dynamics of commodity shipping.}
    \label{fig:graph_representation}
\end{figure}

However, directly using the original graph representation described in Section~\ref{sec:policy_network} presents challenges for the heuristic MCF. Once we get into the MCF phase with the original representation, the graph only includes edges between ports that are already connected by established services from the Network Design Problem (NDP) phase. As shown on the left side of Fig.~\ref{fig:graph_representation}, the edge weights $w$ represent the variable cost of moving an additional FFE of cargo, while edge capacities define the maximum number of FFEs that can flow through those edges. Importantly, in this original representation, $w=0$, since it only accounts for the variable cost of transporting cargo between ports, and it does not capture the internal dynamics of variable costs within a port, such as handling costs.

To effectively implement the heuristic MCF algorithm, the graph representation must be expanded to include these intra-port dynamics. Specifically, the handling costs — such as onloading, offloading, and transshipment — need to be incorporated as edge weights on the graph, allowing for an accurate representation of the network’s cost structure.

Among the variable cost terms, only the handling cost $C_{\text{handle}}$ (as defined in Eq.~\ref{eq:handling_cost}) is considered in this expanded graph representation. The revenue and rejected demand penalty are excluded for the following reasons: revenue is tracked separately within the heuristic MCF algorithm, which will be discussed in detail later, and the rejected demand penalty is uniform across all commodities, making it irrelevant to the heuristic MCF algorithm where the focus is on balancing trade-offs between commodities. Additionally, the fixed costs, as outlined in Eqs.~\ref{eq:service_cost}, \ref{eq:unused_vessel_cost}, and \ref{eq:voyage_cost}, are not included at this stage. These costs are already determined during the Network Design Problem (NDP) phase, and decisions made during the MCF phase will not affect them.

The right side of Fig.~\ref{fig:graph_representation} illustrates this expanded graph representation. A key component of this expansion is the introduction of proxy nodes to represent port-service pairs for each port. For example, if ports $p$ and $q$ are visited by rotations $s_1$, $s_3$ and $s_7$, proxy ports $p_1$, $p_3$ and $p_7$ are created for $p$, and similarly $q_1$, $q_3$ and $q_7$ for $q$.

In the expanded graph, edges connecting $p$ to its proxy nodes $(p, p_s)$, where $s\in \{1,3,7\}$, have weights representing the onloading and offloading costs at port $p$. These edges are assigned infinite capacity, reflecting that there are no logistical constraints on the volume of goods that can be handled at a port. Additionally, edges are introduced between different proxy nodes $p_{s'}$ and $p_{s''}$, where $s'\neq s''$, with infinite capacities. The weights on these edges correspond to transshipment costs at port $p$.

Finally, the original direct connection between ports $p$ and $q$ is now decomposed into edges $(p_s, q_s)$ for each service $s$. The capacity of these edges is determined by the capacity of the respective service, while the weights remain zero, as there is no additional cost for shipping commodities between ports once they are loaded onto a service, as long as the service capacity is not exceeded.

\begin{algorithm}[htp]
\caption{Heuristic MCF with demand revenue prioritization}\label{alg:heuristic_mcf}
\begin{algorithmic}[1]
    \STATE Input: Proposed services $S$, Graph $G(P,E)$, Demand $\mathbf{D}$, Revenue per demand $\mathbf{R}$
    \STATE Initialize: Commodity flows $f_e^d$ = [ ], Missed (rejected) demand $\mathbf{D}_m$ = [ ], Service capacity projected to edges: $\text{Capacity}[e], \forall e \in E$;
    \STATE Descending sort $\mathbf{D}$ based on $\mathbf{R}$;
    \FOR{$d=1,\cdots,D$}
        \STATE Initialize the remaining demand $d_r = d$;
        \STATE Get all available paths $\textbf{T}$ from $d_o$ to $d_d$;
        \STATE Ascending sort $\textbf{T}$ based on \textbf{marginal unit cost} (from $C_{\text{handling}}$);
        \FOR{$t=1,\cdots, \textbf{T}$}
            \STATE Get path capacity: $t[\text{capacity}] = \text{min}(\{\text{Capacity}[e] \text{ for } e \text{ in } t.\text{edges} \})$
            \STATE Flow quantity: $q = \min\{d_r, t[\text{capacity}]\}$
            \STATE $d_r\leftarrow d_r - q$;
            \STATE Append $(d, e, q), \forall e \in t$ to $f_e^d$;
            \STATE Update remaining edge capacities: $\text{Capacity}[e], \forall e \in t$;
            \STATE \textbf{if} $d_r = 0$, break; \textbf{end if}
        \ENDFOR
        \STATE \textbf{if} $d_r > 0$, Append $d_r$ to $\mathbf{D}_m$; \textbf{end if}
    \ENDFOR
    \STATE Return $f_e^d$, $\mathbf{D}_m$
\end{algorithmic}
\end{algorithm}

With the expanded graph representation of the shipping network, we can now develop a greedy sequential commodity flow strategy. For each commodity demand $d$ with origin $d_o$ and destination $d_d$, we use Dijkstra's shortest path algorithm to find the least costly path $t$ between $d_o$ and $d_d$, where the edge weights represent variable costs instead of distances (as illustrated in Fig.~\ref{fig:graph_representation}). Once the cheapest path is identified, we ship a quantity equal to the path's capacity along this route. The path capacity is defined as the minimum capacity among all edges within the path. After shipping, the remaining capacities of all edges in path $t$ are updated to account for the flow of $d$. This process is repeated for any remaining demand of $d$ until no paths with non-zero capacities exist between $d_o$ and $d_d$.

This method is applied sequentially across all commodity demands $\mathbf{D}$, with the order of processing determined by ranking the demands in descending order of their revenue per FFE. The pseudocode for this approach is outlined in Algorithm~\ref{alg:heuristic_mcf}.

After the MCF generates the commodity flow patterns and the corresponding handling costs $C_{\text{handle}}$ (defined in Eq.~\ref{eq:handling_cost}), we calculate the total revenue $R_{\text{total}}$ and the penalties for missed (rejected) demand $C_{\text{reject}}$, using Eqs.~\ref{eq:total_revenue} and \ref{eq:rejected_demand}, respectively.

It is worth mentioning that our implementation of the MCF is written in Rust (\cite{rust}) to ensure high performance. Since the MCF algorithm is executed multiple times during each training iteration, a low-latency solution is essential for maintaining efficiency.

\section{MDP Details}\label{app:mdpdetails}
\begin{figure}[!htp]
    \centering
    \includegraphics[width=0.8\linewidth]{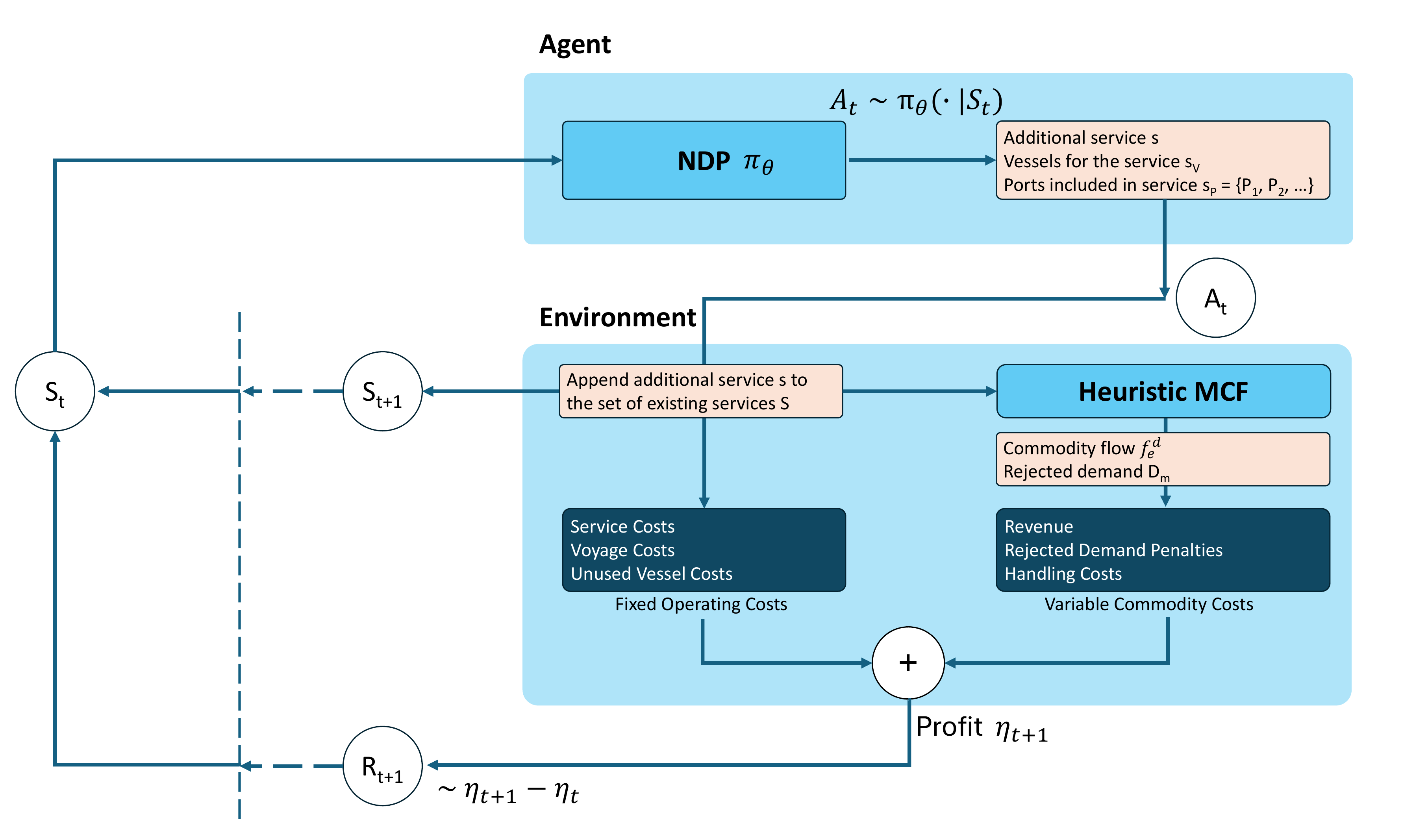}
    \caption{MDP representation of the LSNDP}
    \label{fig:diagram_view}
\end{figure}

Here, we represent the Liner Shipping Network Design Problem (LSNDP) as a Markov Decision Process (MDP). Figure~\ref{fig:diagram_view} provides a high-level overview of this representation.
The action $A_t$ consists of selecting a vessel and determining the sequence of ports for the service $s$, as described in Eq.~\ref{eq:action}. The selected service $s$ is then appended to the existing set of services $S$. This updated set of services is used to directly compute the fixed operating costs. Simultaneously, the updated services are fed into the heuristic MCF algorithm to calculate the variable costs, which include revenue (treated as a negative cost). These fixed and variable costs are summed to compute the total profit $\eta_{t+1}$.

The reward $R_{t+1}$, defined in Eq.\ref{eq:reward}, is then calculated, where both $\eta_{t+1}$ and $\eta_{t}$ are derived from Eq.\ref{eq:mcf_profit} at their respective steps. To enhance the stability of the training process, the rewards at each step are scaled by the initial reward $\eta_1$, as shown in Eq.~\ref{eq:reward_scale} below:
\begin{align}
    R_{t+1} = \frac{\eta_{t+1} - \eta_{t}}{\eta_1}. \label{eq:reward_scale}
\end{align}
The full MDP process can be summarized in Algorithm~\ref{alg:env_step_detailed} below.

Let’s delve deeper into the state representation, denoted by $S_t$. The state at step $t$ is composed of two main components: the vessel state $S_{v,t}$, and the graph state $S_{g,t}$, as defined in Eqs.~\ref{eq:s_g} and \ref{eq:s_v}, respectively. The graph state includes both port features $\mathbf{f}_{p}$ and edge features $\mathbf{f}_{e}$.

\textbf{Port Features:} The port feature vector $\mathbf{f}_{p}$ can be expressed as:
\begin{align}
    \mathbf{f}_p & = [\mathbf{p}_1, \mathbf{p}_2, ...\mathbf{p}_{P+1}] \in \mathbb{R}^{P\times 2},
\end{align}
where each element $\mathbf{p}_{p}$ represents the total incoming and outgoing demand on port $p$:
\begin{align}
    \mathbf{p}_{p} = [\sum_{d_d=p}{d_{q}}, \sum_{d_o=p}{d_{q}}] \in \mathbb{R}^{2}.
\end{align}
Here, the first element of $\mathbf{p}_{p}$ captures the total demand destined for port $p$, and the second element captures the total demand originating from port $p$. Additionally, $P$ represents the total number of ports in the network. The last element, $\mathbf{p}_{P+1}$, represents a global node that remains empty, i.e., $\mathbf{p}_{P+1} = [0,0]$.

\textbf{Edge Features:} The edge feature matrix $\mathbf{f}_{e}$ is composed of both static and dynamic features:
\begin{align}
    \mathbf{f}_e & = 
    \begin{bmatrix}
        \mathbf{f}_e^s \\
        \mathbf{f}_e^d
    \end{bmatrix} \in \mathbb{R}^{D_e\times E},
\end{align}
where $E$ is the number of edges, and $D_e=6+|S|$ is the total number of edge features.

The static features $\mathbf{f}_e^s \in \mathbb{R}^{4\times E}$ consist of four attributes that do not change over time: the origin port index, the destination port index, the distance between the origin and destination, and the revenue generated per unit demand flowing through the edge.

The dynamic features $\mathbf{f}_e^d \in \mathbb{R}^{(2+|S|)\times E}$ include time-varying attributes: the remaining unsatisfied demand between the origin and destination, and the remaining vessel capacity on the edge. Additionally, $|S|$ number of binary indicators track whether an edge is included in service $s$, where a value of 1 indicates inclusion in the service and 0 otherwise.

\textbf{Vessel Features:} The vessel state $\mathbf{v}_t$ represents the characteristics of each vessel class. The dimensionality of the vessel features $D_v = 11$, corresponding to the columns listed in Table 3 of \cite{brouer2014base}, captures key vessel information at each step $t$.


\begin{algorithm}[H]
\caption{Environment Step Function (from step $t$ to $t+1$)}
\label{alg:env_step_detailed}
\SetAlgoLined
\KwIn{Action $A_t$, Previous state $S_{t}$, Remaining vessels $V$, Total demand $D$, Services $S^*$, Profit history $[\eta_0, \cdots, \eta_{t}]$}
\KwOut{Next state $S_{t+1}$, Reward $R_{t+1}$, Done flag}

\textbf{Execute Action and Update State Value:}
\begin{itemize}
    \item Add the action to services, and reduce the remaining vessels:
    \begin{align*}
        (S^*, V) & \gets S_{t} \\
        S^* & \gets S^* + A_t, \\
        V & \gets V - A_t, \\
        S_{t+1} & \gets (S^*, V)
    \end{align*}
\end{itemize}

\textbf{Run MCF Algorithm:}
\begin{itemize}
    \item Execute Algorithm~\ref{alg:heuristic_mcf} with current services $S^*$ on $\mathbf{D}$:
    \[
    f_e^d, \mathbf{D}_m, \text{Capacity}(E) \gets \text{env.mcf}(S^*, \mathbf{D})
    \]
\end{itemize}

\textbf{Compute Cost:}
\begin{itemize}
    \item Calculate the profit using the provided function:
    \[
    \eta_{t+1} \gets \text{env.get\_profit}(S^*, f_e^d, \mathbf{D}_m, V)
    \]
    \item Append the profit to the profit history:
    \[
    [\eta_1, \cdots, \eta_{t}] \gets \eta_{t+1}
    \]
\end{itemize}

\textbf{Check for Termination:}
\begin{itemize}
    \item If all vessel counts are below 0 or remaining demands are 0, terminate:
    \[
    \text{If } (v_n < 0 \ \forall v \in V) \ \text{or} \ (\mathbf{D}_m = 0) \Rightarrow \text{terminate}
    \]
    \item Else, continue the episode
\end{itemize}

\textbf{Calculate Reward:}
\begin{itemize}
    \item Compute the reward based on the change in profit:
    \[
    R_{t+1} \gets \frac{\eta_{t+1} - \eta_{t}}{\eta_1}
    \]
\end{itemize}

\Return{$S_{t+1}$, $R_{t+1}$, \text{Done}}\;
\end{algorithm}

\clearpage

\section{Problem Size and Hyper-parameter Setting}\label{app:hyperparam}
In this section, we summarize the problem size of each LINERLIB instance and the hyper-parameter settings of our approaches in Table~\ref{table:hyperparams}.

\begin{table}[htp]
\centering
\caption{Hyper-parameter settings for encoder / decoder etworks and PPO.}
\label{table:hyperparams}
\begin{tabular}{lll}
\textbf{Hyperparameter} & \textbf{Description} & \textbf{Range of Values} \\
\toprule \toprule
\multicolumn{3}{l}{\textbf{Neural Networks}} \\
Hidden layer         & Number of neurons in the hidden layer       & 512 \\
Transformer head     & Number of attention heads in the transformer & 8 \\
Transformer layer    & Number of transformer layers               & 3 \\
GAT layer            & Number of Graph Attention Network layers      & 3 \\
LSTM layer           & Number of LSTM layers                      & 1 \\
\midrule
\multicolumn{3}{l}{\textbf{PPO}} \\
Learning rate        & Constant learning rate for optimizer                 & [1e-4, 3e-4] \\
Environment          & Number of parallel environments            & [8, 16] \\
Step                 & Number of steps per environment per update & [50, 100] \\
Discount factor      & Discount factor for future rewards         & 1 \\
TD lambda            & Lambda parameter for TD learning           & 0.9 \\
Mini-batch size      & Size of mini-batches for training          & [64, 128] \\
Update epoch         & Number of epochs per update                & 10 \\
Clip coefficient     & Clipping coefficient for PPO               & [0.15, 0.25] \\
Target KL            & Target Kullback-Leibler divergence         & 0.1 \\
Entropy coefficient  & Coefficient for entropy regularization     & [0.01, 0.1] \\
Value function coef  & Coefficient for value function loss        & 0.5 \\
\bottomrule \bottomrule
\end{tabular}
\label{table:hyperparameters}
\end{table}

We use the \textbf{AdamW} \footnote{https://pytorch.org/docs/stable/generated/torch.optim.AdamW.html} optimizer on Pytorch, with default values for $\beta_1, \beta_2$ and weight decay. Our environment is implemented with the \textbf{Gymnasium} Interface\footnote{https://github.com/Farama-Foundation/Gymnasium}.

\clearpage
\section{Perturbation Profiles}\label{app:perturb}
\begin{figure}[htp]
    \centering
    \includegraphics[width=0.8\linewidth]{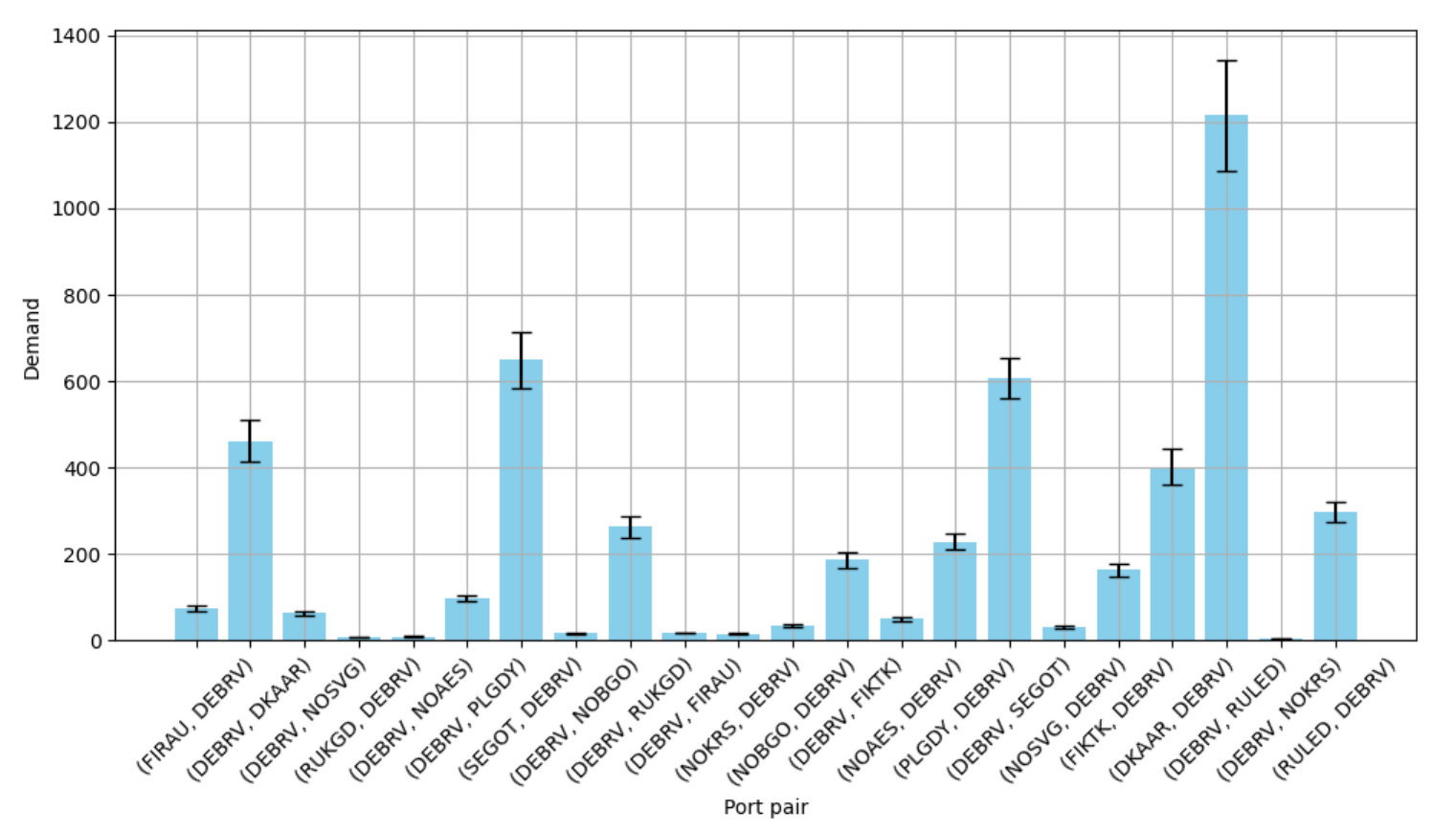}
    \caption{Perturbed demand quantities (in FFEs) for the Baltic instance with a perturbation level of 10\%.}
    \label{fig:perturb1}
\end{figure}

\begin{figure}[htp]
    \centering
    \includegraphics[width=0.8\linewidth]{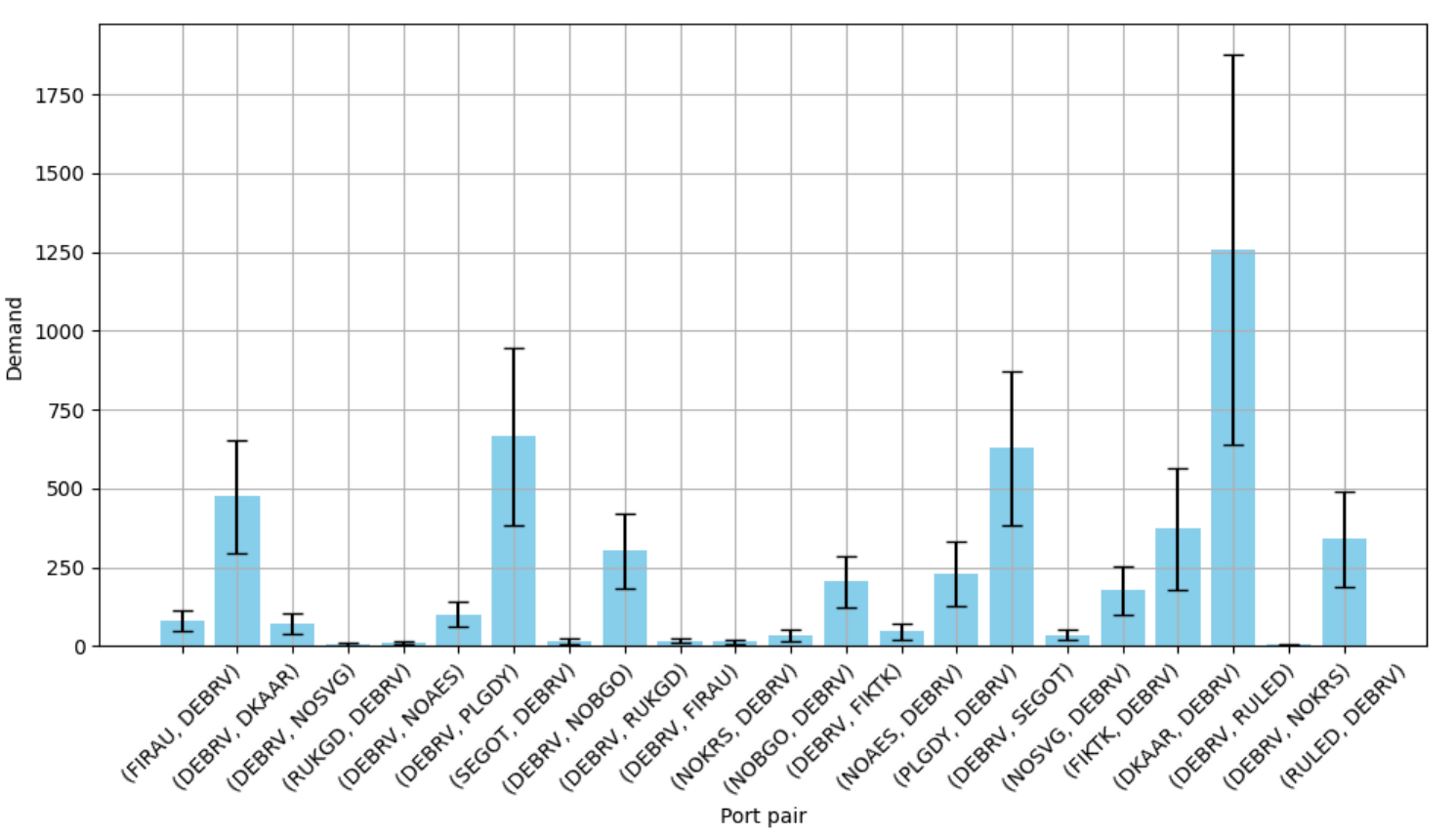}
    \caption{Perturbed demand quantities (in FFEs) for the Baltic instance with a perturbation level of 50\%.}
    \label{fig:perturb2}
\end{figure}

In this section, we discuss the perturbations applied to the demand quantities in the Baltic instance. Figures~\ref{fig:perturb1} and \ref{fig:perturb2} illustrate the $\pm 10\%$ and $\pm 50\%$ perturbations, respectively. In both figures, the blue bars represent the original demand quantities from the Baltic instance, while the black error bars indicate the range of perturbation. The x-axis shows the origin and destination ports for each demand.

The 10\% and 50\% perturbations correspond to the standard deviation of the distribution used to generate new demand values. For example, when generating a perturbed instance with 10\% perturbation, the new demand quantities are randomly sampled from a distribution where the original demand $d_q$ is the mean, and the standard deviation is $10\% \cdot d_q$.

Additionally, note that the samples are truncated at 0 to prevent the generation of instances with negative demand values.

\clearpage
\section{Network Design Comparison for the Baltic Instance}\label{app:baltic}
\begin{figure}[htp]
    \centering
    \includegraphics[width=0.65\linewidth]{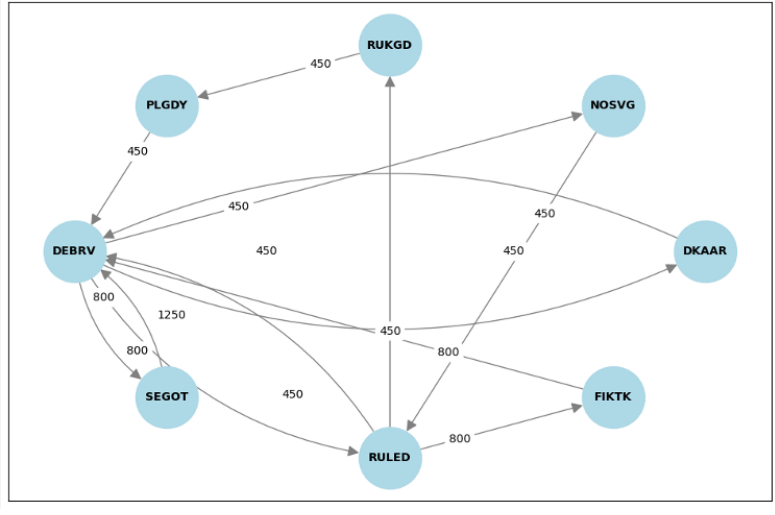}
    \caption{Network design for Baltic instance obtained by the RL-based approach. Ports are connected by vessels with specified capacity. The design uses 2.03 vessels of capacity 800 and 4.31 vessels of capacity 450. The total net profit is \$276,428.}
    \label{fig:design1}
\end{figure}

Figure~\ref{fig:design1} illustrates the network design for the Baltic instance generated by the RL agent using the encoder-decoder approach. This design corresponds to the experimental setup described in Section~\ref{subsec:baltic}, where training and validation were performed on a dataset comprising 16,000 instances, with demand quantities perturbed by $\pm 10\%$ from the original LINERLIB Baltic instance.

\begin{figure}[htp]
    \centering
    \includegraphics[width=0.65\linewidth]{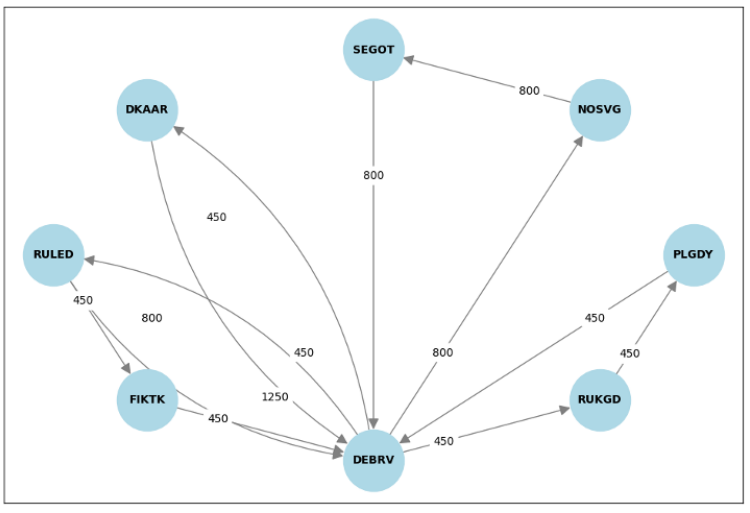}
    \caption{Network design for Baltic instance obtained by the LINERLIB solution. Ports are connected by vessels with specified capacity. The design uses 2.14 vessels of capacity 800 and 3.58 vessels of capacity 450. The total net profit is \$260,948.}
    \label{fig:design2}
\end{figure}

Figure~\ref{fig:design2} shows the network design for the Baltic instance produced by the LINERLIB solution. A comparison of the two figures reveals that the RL-based design covers the same 8 out of 12 ports as the LINERLIB solution. Additionally, both designs propose 5 simple services that follow a similar pattern.

\clearpage
\section{Network Design Comparison for the WAF and World Small instances}\label{app:waf_ws}
\begin{figure}[htp]
    \centering
    \includegraphics[width=0.7\linewidth]{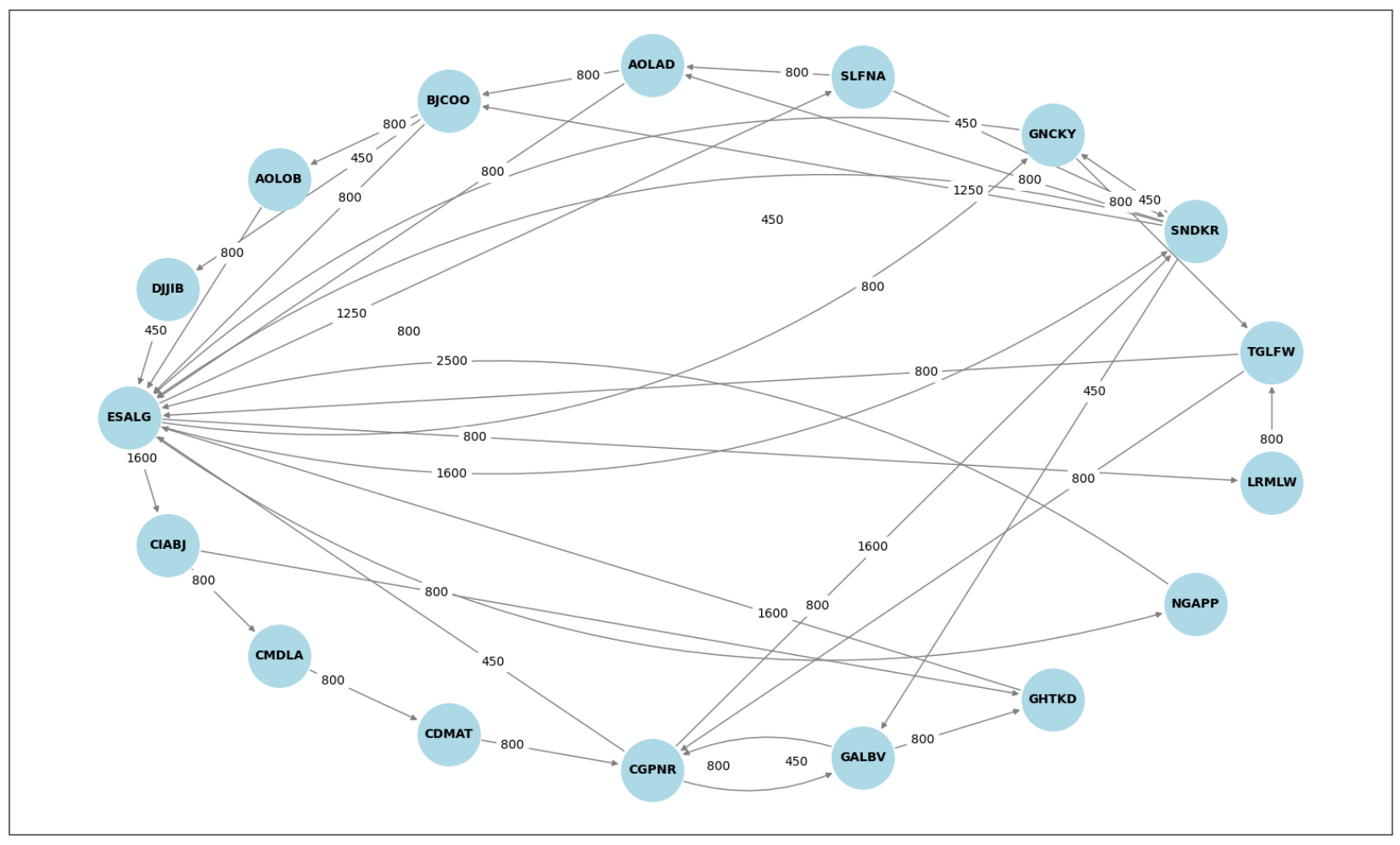}
    \caption{Network design for WAF instance obtained by running our encoder-decoder approach. Ports are connected by vessel with specified capacity. It uses 31.68 vessels of capacity 800 and 14.05 vessels of capacity 450. The total net profit is \$5,596,382.}
    \label{fig:design3}
\end{figure}

\begin{figure}[htp]
    \centering
    \includegraphics[width=0.7\linewidth]{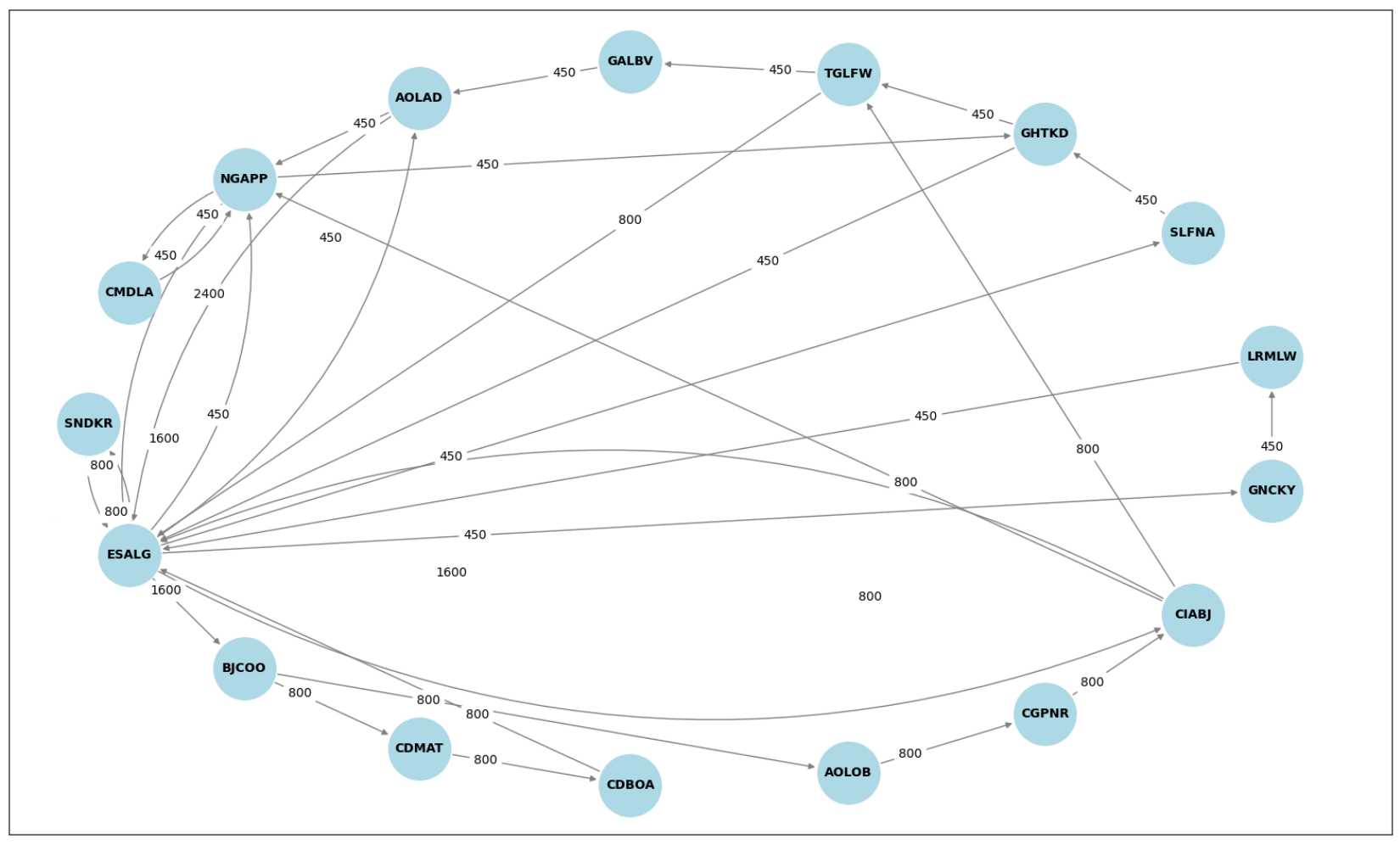}
    \caption{Network design for WAF instance obtained by the LINERLIB solution. Ports are connected by vessel with specified capacity. It uses 21.34 vessels of capacity 800 and 12.80 vessels of capacity 450. The total net profit is \$5,202,534.}
    \label{fig:design4}
\end{figure}

In this section, we utilize the RL-based NDP solution as an optimizer, where the RL agent's training process functions similarly to a traditional optimization solver, eliminating the distinction between training and inference phases (as outlined in Section~\ref{subsec:optimizer}). We compare the network designs generated by our encoder-decoder approach and the LINERLIB solution for the WAF instance in Figure~\ref{fig:design3} and Figure~\ref{fig:design4}, respectively. Additionally, the schedule outputs for the World Small instance from both our encoder-decoder approach and the LINERLIB solution are compared in Table~\ref{table:design5} and Table~\ref{table:design6_1}. Finally, we compare vessel usage between the two approaches for the World Small instance in Table~\ref{table:vessel}. Even though we obtain different network designs from the LINERLIB solution, the total net profits of our RL-based approach for both WAF and WorldSmall instances are higher than the LINERLIB benchmark, which shows the capability of our RL-based approach being used as an optimizer.

\clearpage
\begin{table}[htp]
\centering
\caption{Schedule output for World Small instance obtained by our encoder-decoder approach.}
\begin{tabular}{llSS}

            & Schedule & {Vessel capacity} & {Number of vessels} 
\\ \toprule \toprule
Schedule 1 & \makecell[l]{CNYTN $\rightarrow$ ITGIT $\rightarrow$ BEANR $\rightarrow$ PAMIT $\rightarrow$ USMIA \\ $\rightarrow$ NGAPP $\rightarrow$ HKHKG $\rightarrow$ MAPTM $\rightarrow$ CNYTN} & 450 & 24.11 
\\ \hline
Schedule 2 & \makecell[l]{CNYTN $\rightarrow$ TWKHH $\rightarrow$ NZAKL $\rightarrow$ CNSHA $\rightarrow$ MYTPP \\ $\rightarrow$ AEJEA $\rightarrow$ INNSA $\rightarrow$ SAJED $\rightarrow$ ZADUR $\rightarrow$ KEMBA \\ $\rightarrow$ HKHKG $\rightarrow$ LKCMB $\rightarrow$ OMSLL $\rightarrow$ MYPKG $\rightarrow$ CNYTN} & 800 & 19.29 
\\ \hline
Schedule 3 & \makecell[l]{CNYTN $\rightarrow$ TWKHH $\rightarrow$ NZAKL $\rightarrow$ PABLB $\rightarrow$ CNSHA \\ $\rightarrow$ MYTPP $\rightarrow$ LKCMB $\rightarrow$ PKBQM $\rightarrow$ HKHKG $\rightarrow$ MYPKG \\ $\rightarrow$ KEMBA $\rightarrow$ INNSA $\rightarrow$ SGSIN $\rightarrow$ CNYTN} & 800 & 19.81
\\ \hline
Schedule 4 & \makecell[l]{CNYTN $\rightarrow$ TWKHH $\rightarrow$ NZAKL $\rightarrow$ PABLB $\rightarrow$ CLSAI \\ $\rightarrow$ JPYOK $\rightarrow$ CNSHA $\rightarrow$ AEJEA $\rightarrow$ PKBQM $\rightarrow$ KEMBA \\ $\rightarrow$ ZADUR $\rightarrow$ OMSLL $\rightarrow$ SGSIN $\rightarrow$ LKCMB $\rightarrow$ MYPKG \\ $\rightarrow$ INNSA $\rightarrow$ AOLAD $\rightarrow$ MYTPP $\rightarrow$ HKHKG $\rightarrow$ SAJED \\ $\rightarrow$ TRAMB $\rightarrow$ CNYTN} & 1200 & 28.98
\\ \hline
Schedule 5 & \makecell[l]{CNYTN $\rightarrow$ TWKHH $\rightarrow$ NZAKL $\rightarrow$ CLSAI $\rightarrow$ PABLB \\ $\rightarrow$ ECGYE $\rightarrow$ PAMIT $\rightarrow$ USEWR $\rightarrow$ USMIA $\rightarrow$ BEZEE \\ $\rightarrow$ BEANR $\rightarrow$ NLRTM $\rightarrow$ DEHAM $\rightarrow$ ITGIT $\rightarrow$ EGPSD \\ $\rightarrow$ MAPTM $\rightarrow$ ESBCN $\rightarrow$ CAMTR $\rightarrow$ DEBRV $\rightarrow$ GBFXT \\ $\rightarrow$ TRAMB $\rightarrow$ SAJED $\rightarrow$ INNSA $\rightarrow$ SGSIN $\rightarrow$ AEJEA \\ $\rightarrow$ LKCMB $\rightarrow$ OMSLL $\rightarrow$ KEMBA $\rightarrow$ AOLAD $\rightarrow$ GHTKD \\ $\rightarrow$ UYMVD $\rightarrow$ MYTPP $\rightarrow$ MYPKG $\rightarrow$ PKBQM $\rightarrow$ ZADUR \\ $\rightarrow$ HKHKG $\rightarrow$ NGAPP $\rightarrow$ CNYTN} & 1200 & 41.42
\\ \hline
Schedule 6 & \makecell[l]{CNYTN $\rightarrow$ HKHKG $\rightarrow$ DEBRV $\rightarrow$ NLRTM $\rightarrow$ GBFXT \\ $\rightarrow$ USCHS $\rightarrow$ ITGIT $\rightarrow$ ESALG $\rightarrow$ NGAPP $\rightarrow$ BRSSZ \\ $\rightarrow$ ZADUR $\rightarrow$ AEJEA $\rightarrow$ MYTPP $\rightarrow$ CNSHA $\rightarrow$ CNYTN} & 2400 & 17.56
\\ \hline
Schedule 7 & \makecell[l]{CNYTN $\rightarrow$ HKHKG $\rightarrow$ CNTAO $\rightarrow$ CNSHA $\rightarrow$ ITGIT \\ $\rightarrow$ EGPSD $\rightarrow$ MYTPP $\rightarrow$ CNYTN} & 2400 & 7.38
\\ \hline
Schedule 8 & \makecell[l]{CNYTN $\rightarrow$ HKHKG $\rightarrow$ JPYOK $\rightarrow$ CNSHA $\rightarrow$ MYTPP \\ $\rightarrow$ ITGIT $\rightarrow$ EGPSD $\rightarrow$ SAJED $\rightarrow$ LKCMB $\rightarrow$ CNYTN} & 2400 & 8.12
\\ \hline
Schedule 9 & \makecell[l]{CNYTN $\rightarrow$ HKHKG $\rightarrow$ JPYOK $\rightarrow$ KRPUS $\rightarrow$ CNSHA \\ $\rightarrow$ MYTPP $\rightarrow$ ITGIT $\rightarrow$ EGPSD $\rightarrow$ SAJED $\rightarrow$ AEJEA \\ $\rightarrow$ INNSA $\rightarrow$ LKCMB $\rightarrow$ CNYTN} & 2400 & 9.18
\\ \hline
Schedule 10 & \makecell[l]{CNYTN $\rightarrow$ HKHKG $\rightarrow$ CNSHA $\rightarrow$ JPYOK $\rightarrow$ USLAX \\ $\rightarrow$ CNTAO $\rightarrow$ KRPUS $\rightarrow$ TWKHH $\rightarrow$ CNYTN} & 2400 & 6.45
\\ \hline
Schedule 11 & \makecell[l]{CNYTN $\rightarrow$ HKHKG $\rightarrow$ CNTAO $\rightarrow$ KRPUS $\rightarrow$ USLAX \\ $\rightarrow$ CAVAN $\rightarrow$ JPYOK $\rightarrow$ CNSHA $\rightarrow$ TWKHH $\rightarrow$ NGAPP \\ $\rightarrow$ GHTKD $\rightarrow$ CNYTN} & 2400 & 14.10
\\ \hline
Schedule 12 & \makecell[l]{CNYTN $\rightarrow$ HKHKG $\rightarrow$ CNTAO $\rightarrow$ CNSHA $\rightarrow$ USLAX \\ $\rightarrow$ CAVAN $\rightarrow$ JPYOK $\rightarrow$ KRPUS $\rightarrow$ TWKHH $\rightarrow$ AUBNE \\ $\rightarrow$ CNYTN} & 2400 & 9.80
\\ \hline
Schedule 13 & \makecell[l]{CNYTN $\rightarrow$ HKHKG $\rightarrow$ TWKHH $\rightarrow$ CNSHA $\rightarrow$ CNTAO \\ $\rightarrow$ AEJEA $\rightarrow$ INNSA $\rightarrow$ OMSLL $\rightarrow$ MYTPP $\rightarrow$ CNYTN} & 2400 & 6.60
\\ \hline
Schedule 14 & \makecell[l]{CNYTN $\rightarrow$ MYTPP $\rightarrow$ ESALG $\rightarrow$ GBFXT $\rightarrow$ DEBRV \\ $\rightarrow$ NLRTM $\rightarrow$ NGAPP $\rightarrow$ ZADUR $\rightarrow$ HKHKG $\rightarrow$ TRAMB \\ $\rightarrow$ BEANR $\rightarrow$ BEZEE $\rightarrow$ ITGIT $\rightarrow$ SAJED $\rightarrow$ LKCMB \\ $\rightarrow$ CNYTN} & 4200 & 18.23
\\ \hline
Schedule 15 & \makecell[l]{CNYTN $\rightarrow$ GHTKD $\rightarrow$ USCHS $\rightarrow$ PAMIT $\rightarrow$ USMIA \\ $\rightarrow$ NLRTM $\rightarrow$ AEJEA $\rightarrow$ MYTPP $\rightarrow$ ITGIT $\rightarrow$ GBFXT \\ $\rightarrow$ DEBRV $\rightarrow$ MAPTM $\rightarrow$ CNYTN} & 4200 & 19.41
\\ \hline
Schedule 16 & \makecell[l]{CNYTN $\rightarrow$ ZADUR $\rightarrow$ USCHS $\rightarrow$ PAMIT $\rightarrow$ USMIA \\ $\rightarrow$ USEWR $\rightarrow$ BRSSZ $\rightarrow$ DEBRV $\rightarrow$ ESALG $\rightarrow$ GHTKD \\ $\rightarrow$ NGAPP $\rightarrow$ CNYTN} & 4200 & 17.84
\\ \hline
Schedule 17 & \makecell[l]{CNYTN $\rightarrow$ CNSHA $\rightarrow$ MYTPP $\rightarrow$ CNYTN} & 4200 & 2.07
\\ \hline
Schedule 18 & \makecell[l]{CNYTN $\rightarrow$ CNSHA $\rightarrow$ CNYTN} & 4200 & 0.88
\\ \hline
Schedule 19 & \makecell[l]{CNYTN $\rightarrow$ CNTAO $\rightarrow$ CNSHA $\rightarrow$ ZADUR $\rightarrow$ NGAPP \\ $\rightarrow$ NLRTM $\rightarrow$ DEHAM $\rightarrow$ DEBRV $\rightarrow$ BEANR $\rightarrow$ GBFXT \\ $\rightarrow$ EGPSD $\rightarrow$ MYTPP $\rightarrow$ CNYTN} & 7500 & 11.07
\\ \bottomrule \bottomrule
\end{tabular}%
\label{table:design5}
\end{table}

\begin{table}[htp]
\centering
\caption{Schedule output for World Small instance obtained by the LINERLIB solution.}
\begin{tabular}{llSS}

            & Schedule & {Vessel capacity} & {Number of vessels} 
\\ \toprule \toprule
Schedule 1 & NGAPP $\rightarrow$ GHTKD $\rightarrow$ NGAPP & 450 & 0.60 
\\ \hline
Schedule 2 & ZADUR $\rightarrow$ AOLAD $\rightarrow$ MAPTM $\rightarrow$ GHTKD $\rightarrow$ ZADUR & 450 & 6.68 
\\ \hline
Schedule 3 & \makecell[l]{OMSLL $\rightarrow$ SAJED $\rightarrow$ KEMBA $\rightarrow$ LKCMB $\rightarrow$ INNSA \\ $\rightarrow$ OMSLL} & 450 & 4.73 
\\ \hline
Schedule 4 & AEJEA $\rightarrow$ INNSA $\rightarrow$ AEJEA & 450 & 1.46 
\\ \hline
Schedule 5 & SAJED $\rightarrow$ KEMBA $\rightarrow$ SAJED & 450 & 2.57 
\\ \hline
Schedule 6 & INNSA $\rightarrow$ LKCMB $\rightarrow$ KEMBA $\rightarrow$ INNSA & 450 & 3.31 
\\ \hline
Schedule 7 & PABLB $\rightarrow$ ECGYE $\rightarrow$ PABLB & 450 & 1.10 
\\ \hline
Schedule 8 & PAMIT $\rightarrow$ USCHS $\rightarrow$ ECGYE $\rightarrow$ PAMIT & 450 & 2.97 
\\ \hline
Schedule 9 & \makecell[l]{LKCMB $\rightarrow$ MYTPP $\rightarrow$ TWKHH $\rightarrow$ CNTAO $\rightarrow$ HKHKG \\ $\rightarrow$ MYPKG  $\rightarrow$ INNSA $\rightarrow$ LKCMB} & 800 & 5.33
\\ \hline
Schedule 10 & PKBQM $\rightarrow$ SAJED $\rightarrow$ OMSLL $\rightarrow$ INNSA $\rightarrow$ PKBQM & 800 & 2.73 
\\ \hline
Schedule 11 & \makecell[l]{MAPTM $\rightarrow$ USEWR $\rightarrow$ USLAX $\rightarrow$ PABLB $\rightarrow$ PAMIT \\ $\rightarrow$ BRSSZ $\rightarrow$ NGAPP $\rightarrow$ ESALG $\rightarrow$ MAPTM} & 800 & 12.15
\\ \hline
Schedule 12 & ITGIT $\rightarrow$ ESALG $\rightarrow$ ITGIT & 800 & 1.18 
\\ \hline
Schedule 13 & EGPSD $\rightarrow$ TRAMB $\rightarrow$ EGPSD & 800 & 0.99 
\\ \hline
Schedule 14 & ITGIT $\rightarrow$ TRAMB $\rightarrow$ ITGIT & 800 & 1.15 
\\ \hline
Schedule 15 & PKBQM $\rightarrow$ LKCMB $\rightarrow$ AEJEA $\rightarrow$ PKBQM & 800 & 2.55 
\\ \hline
Schedule 16 & INNSA $\rightarrow$ AEJEA $\rightarrow$ INNSA & 800 & 1.30 
\\ \hline
Schedule 17 & \makecell[l]{JPYOK $\rightarrow$ KRPUS $\rightarrow$ CNTAO $\rightarrow$ CAVAN $\rightarrow$ SGSIN \\ $\rightarrow$ MYTPP $\rightarrow$ AEJEA $\rightarrow$ SAJED $\rightarrow$ OMSLL \\ $\rightarrow$ MYPKG $\rightarrow$ HKHKG $\rightarrow$ CNYTN $\rightarrow$ TWKHH \\ $\rightarrow$ CNSHA $\rightarrow$ JPYOK} & 1200 & 11.06
\\ \hline
Schedule 18 & \makecell[l]{EGPSD $\rightarrow$ SAJED $\rightarrow$ MYPKG $\rightarrow$ HKHKG $\rightarrow$ MYTPP \\ $\rightarrow$ LKCMB $\rightarrow$ OMSLL $\rightarrow$ ESBCN $\rightarrow$ ITGIT $\rightarrow$ EGPSD} & 1200 & 6.75
\\ \hline
Schedule 19 & CNSHA $\rightarrow$ JPYOK $\rightarrow$ CAVAN $\rightarrow$ CNTAO $\rightarrow$ CNSHA & 1200 & 4.15
\\ \hline
Schedule 20 & MYPKG $\rightarrow$ EGPSD $\rightarrow$ OMSLL $\rightarrow$ MYTPP $\rightarrow$ MYPKG & 1200 & 3.96
\\ \hline
Schedule 21 & \makecell[l]{SGSIN $\rightarrow$ HKHKG $\rightarrow$ KRPUS $\rightarrow$ CNTAO $\rightarrow$ USLAX \\ $\rightarrow$ CNSHA $\rightarrow$ TWKHH $\rightarrow$ MYTPP $\rightarrow$ SGSIN} & 1200 & 6.72
\\ \hline
Schedule 22 & \makecell[l]{MYPKG $\rightarrow$ MYTPP $\rightarrow$ TWKHH $\rightarrow$ CAVAN $\rightarrow$ AUBNE \\ $\rightarrow$ MYPKG} & 1200 & 6.71
\\ \hline
Schedule 23 & NZAKL $\rightarrow$ AUBNE $\rightarrow$ NZAKL & 1200 & 1.17
\\ \hline
Schedule 24 & \makecell[l]{KRPUS $\rightarrow$ MYPKG $\rightarrow$ MYTPP $\rightarrow$ AUBNE $\rightarrow$ NZAKL \\ $\rightarrow$ USLAX  $\rightarrow$ CAVAN $\rightarrow$ JPYOK $\rightarrow$ TWKHH \\ $\rightarrow$ HKHKG $\rightarrow$ CNTAO $\rightarrow$ KRPUS} & 1200 & 9.06
\\ \hline
Schedule 25 & \makecell[l]{USEWR $\rightarrow$ USCHS $\rightarrow$ GBFXT $\rightarrow$ DEBRV $\rightarrow$ ESALG \\ $\rightarrow$ MAPTM $\rightarrow$ EGPSD $\rightarrow$ SAJED $\rightarrow$ USEWR} & 1200 & 8.15
\\ \hline
Schedule 26 & ITGIT $\rightarrow$ SAJED $\rightarrow$ ITGIT & 1200 & 1.41
\\ \hline
Schedule 27 & \makecell[l]{MYTPP $\rightarrow$ CNYTN $\rightarrow$ OMSLL $\rightarrow$ SAJED $\rightarrow$ NGAPP \\ $\rightarrow$ ZADUR $\rightarrow$ CNSHA $\rightarrow$ MYTPP} & 2400 & 10.54
\\ \hline
Schedule 28 & LKCMB $\rightarrow$ ZADUR $\rightarrow$ SAJED $\rightarrow$ AEJEA $\rightarrow$ LKCMB & 2400 & 4.97
\\ \hline
Schedule 29 & \makecell[l]{AUBNE $\rightarrow$ NZAKL $\rightarrow$ TWKHH $\rightarrow$ CNYTN $\rightarrow$ SGSIN \\ $\rightarrow$ AUBNE} & 2400 & 5.25
\\ \hline
Schedule 30 & MYTPP $\rightarrow$ SGSIN $\rightarrow$ MYTPP & 2400 & 0.31
\\ \bottomrule \bottomrule
\end{tabular}%
\label{table:design6_1}
\end{table}

\addtocounter{table}{-1}

\begin{table}[htp]
\centering
\caption{Schedule output for World Small instance obtained by the LINERLIB solution (cont.).}
\begin{tabular}{llSS}

            & Schedule & {Vessel capacity} & {Number of vessels} 
\\ \toprule \toprule
Schedule 31 & \makecell[l]{PABLB $\rightarrow$ USEWR $\rightarrow$ PAMIT $\rightarrow$ ESALG $\rightarrow$ NLRTM \\ $\rightarrow$ USCHS $\rightarrow$ USMIA $\rightarrow$ CLSAI $\rightarrow$ PABLB} & 2400 & 8.92
\\ \hline
Schedule 32 & USCHS $\rightarrow$ PAMIT $\rightarrow$ USCHS & 2400 & 1.11
\\ \hline
Schedule 33 & \makecell[l]{CLSAI $\rightarrow$ USLAX $\rightarrow$ CAVAN $\rightarrow$ PABLB $\rightarrow$ PAMIT \\ $\rightarrow$ CLSAI} & 2400 & 6.04
\\ \hline
Schedule 34 & \makecell[l]{ITGIT $\rightarrow$ MAPTM $\rightarrow$ BRSSZ $\rightarrow$ GBFXT $\rightarrow$ DEBRV \\ $\rightarrow$ SAJED $\rightarrow$ EGPSD $\rightarrow$ ITGIT} & 2400 & 7.36
\\ \hline
Schedule 35 & \makecell[l]{BEANR $\rightarrow$ DEBRV $\rightarrow$ DEHAM $\rightarrow$ ITGIT $\rightarrow$ MAPTM \\ $\rightarrow$ BEANR} & 2400 & 2.79
\\ \hline
Schedule 36 & \makecell[l]{USMIA $\rightarrow$ NLRTM $\rightarrow$ BEANR $\rightarrow$ EGPSD $\rightarrow$ SAJED \\ $\rightarrow$ AEJEA $\rightarrow$ NGAPP $\rightarrow$ PAMIT $\rightarrow$ USMIA} & 2400 & 9.99
\\ \hline
Schedule 37 & \makecell[l]{CNYTN $\rightarrow$ HKHKG $\rightarrow$ SGSIN $\rightarrow$ MYTPP $\rightarrow$ MYPKG \\ $\rightarrow$ ZADUR $\rightarrow$ CNTAO $\rightarrow$ CNYTN} & 2400 & 6.68
\\ \hline
Schedule 38 & \makecell[l]{BRSSZ $\rightarrow$ PAMIT $\rightarrow$ USMIA $\rightarrow$ USCHS $\rightarrow$ ESALG \\ $\rightarrow$ ITGIT $\rightarrow$ MAPTM $\rightarrow$ NGAPP $\rightarrow$ BRSSZ} & 2400 & 7.72
\\ \hline
Schedule 39 & \makecell[l]{ITGIT $\rightarrow$ AEJEA $\rightarrow$ MYTPP $\rightarrow$ OMSLL $\rightarrow$ SAJED \\ $\rightarrow$ BEZEE $\rightarrow$ NLRTM $\rightarrow$ MAPTM $\rightarrow$ ITGIT} & 4200 & 7.83
\\ \hline
Schedule 40 & \makecell[l]{ESBCN $\rightarrow$ GBFXT $\rightarrow$ NLRTM $\rightarrow$ DEBRV \\ $\rightarrow$ EGPSD $\rightarrow$ SAJED $\rightarrow$ ESBCN} & 4200 & 4.06
\\ \hline
Schedule 41 & LKCMB $\rightarrow$ SAJED $\rightarrow$ LKCMB & 4200 & 2.30
\\ \hline
Schedule 42 & MYTPP $\rightarrow$ MYPKG $\rightarrow$ LKCMB $\rightarrow$ MYTPP & 4200 & 1.67
\\ \hline
Schedule 43 & \makecell[l]{AEJEA $\rightarrow$ SAJED $\rightarrow$ ITGIT $\rightarrow$ ESALG $\rightarrow$ LKCMB \\ $\rightarrow$ AEJEA} & 4200 & 5.19
\\ \hline
Schedule 44 & \makecell[l]{HKHKG $\rightarrow$ CNYTN $\rightarrow$ KRPUS $\rightarrow$ CAVAN \\ $\rightarrow$ USLAX $\rightarrow$ JPYOK $\rightarrow$ MYTPP $\rightarrow$ HKHKG} & 4200 & 6.85
\\ \hline
Schedule 45 & LKCMB $\rightarrow$ MYTPP $\rightarrow$ LKCMB & 4200 & 1.45
\\ \hline
Schedule 46 & \makecell[l]{OMSLL $\rightarrow$ SGSIN $\rightarrow$ MYTPP $\rightarrow$ HKHKG $\rightarrow$ CNSHA \\ $\rightarrow$ MYPKG $\rightarrow$ AEJEA $\rightarrow$ EGPSD $\rightarrow$ ITGIT $\rightarrow$ OMSLL} & 4200 & 7.86
\\ \hline
Schedule 47 & GBFXT $\rightarrow$ NLRTM $\rightarrow$ ITGIT $\rightarrow$ GBFXT & 4200 & 2.20
\\ \hline
Schedule 48 & \makecell[l]{HKHKG $\rightarrow$ SGSIN $\rightarrow$ MYPKG $\rightarrow$ AEJEA $\rightarrow$ SAJED \\ $\rightarrow$ OMSLL $\rightarrow$ MYTPP $\rightarrow$ JPYOK $\rightarrow$ KRPUS $\rightarrow$ CNSHA \\ $\rightarrow$ HKHKG} & 4200 & 7.48
\\ \hline
Schedule 49 & OMSLL $\rightarrow$ ESALG $\rightarrow$ DEBRV $\rightarrow$ SAJED $\rightarrow$ OMSLL & 4200 & 4.57
\\ \hline
Schedule 50 & \makecell[l]{GBFXT $\rightarrow$ USCHS $\rightarrow$ CAMTR $\rightarrow$ DEBRV $\rightarrow$ NLRTM \\ $\rightarrow$ GBFXT} & 7500 & 3.94
\\ \hline
Schedule 51 & CNYTN $\rightarrow$ ITGIT $\rightarrow$ HKHKG $\rightarrow$ CNYTN & 7500 & 5.61
\\ \bottomrule \bottomrule
\end{tabular}%
\end{table}

\begin{table}[htp]
\centering
\caption{Comparison of Vessel usage from our RL-based encoder-decoder approach and the LINERLIB solution for the World Small instance.}
\begin{tabular}{lSS}

                  & {RL-based} & {LINERLIB solution} 
                  \\ \toprule \toprule
Vessel capacity 450   & 24.11  & 23.43 \\ \hline
Vessel capacity 800 & 39.10 & 27.37 \\ \hline
Vessel capacity 1200 & 70.40    & 59.14 \\ \hline
Vessel capacity 2400 & 79.19   & 71.68 \\ \hline
Vessel capacity 4200 & 58.43 & 51.47 \\ \hline
Vessel capacity 7500 & 11.07    & 9.55 \\ \hline
Total vessel usage  & 282.30 & 242.64
\\ \bottomrule \bottomrule
\end{tabular}%
\label{table:vessel}
\end{table}

\end{document}